%% file: main.tex
\PassOptionsToPackage{square,comma,numbers,sort&compress}{natbib}
\documentclass[10pt,twocolumn,letterpaper]{article}

\usepackage{cvpr}              
\usepackage{cite}
\usepackage{hwemoji}
\usepackage{amssymb}

\usepackage{textcomp}
\usepackage{mdframed}
\usepackage{amsmath}
\usepackage{CJKutf8}
\usepackage{algorithm} 
\usepackage{algpseudocode}
\usepackage{framed}
\usepackage{xcolor}
\usepackage{listings}
\usepackage{soul}
\usepackage{amssymb}
\usepackage{tcolorbox}

\usepackage{soul}
\usepackage{amssymb}
\usepackage{multirow}
\definecolor{cvprblue}{rgb}{0.21,0.49,0.74}
\definecolor{myblue}{rgb}{0.97,0.53,0.53}
\definecolor{logo}{rgb}{0.694,0.372,0.145}
\definecolor{logo2}{rgb}{0.772,0.403,0.1412} 
\definecolor{myred}{rgb}{0.1,0.53,0.99}
\definecolor{hld}{rgb}{0.97,0.81,0.80} 
\definecolor{hlg}{rgb}{1.0,0.90,0.8} 
\definecolor{highlightcolor}{RGB}{255,255,204}  
\definecolor{prompt}{rgb}{0.85,0.91,0.99} 
\definecolor{image}{rgb}{0.8,0.99,0.8} 
\definecolor{condition}{rgb}{1.0,0.8,0.901} 
\definecolor{response}{rgb}{0.88,0.835,0.906} 

\usepackage[pagebackref,breaklinks,colorlinks,citecolor=cvprblue]{hyperref}

\def\paperID{2286} 
\def\confName{CVPR}
\def\confYear{2024}
 
\newcommand{\Sm}{\boldsymbol{\mathcal{S}}}
\newcommand{\Imm}{\boldsymbol{I}}
\newcommand{\Rm}{\boldsymbol{R}}
\newcommand{\Cm}{\boldsymbol{C}} 
\title{\includegraphics[width=0.044\textwidth]{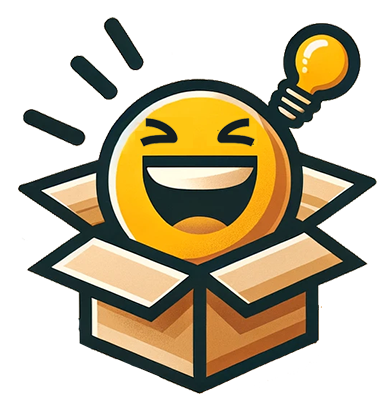} Let's Think Outside the Box: Exploring Leap-of-Thought in Large Language Models with Creative Humor Generation}

\author{Shanshan Zhong$^{1,2}$\thanks{co-first author \quad $\dagger$ corresponding author} \quad Zhongzhan Huang$^{1,2*}$ \quad Shanghua Gao$^3$ \quad Wushao Wen$^2$  \quad Liang Lin$^{2}$\\ \quad Marinka Zitnik$^{3}$ \quad Pan Zhou$^{1\dagger}$\\
$^1$Sea AI Lab\quad $^2$Sun Yat-sen University \quad $^3$Harvard University \\
{\tt\small $^*$Co-first author:~\{zhongshsh5,huangzhzh23\}@mail2.sysu.edu.cn }\\
{\tt\small  $^\dagger$Corresponding author:~zhoupan@sea.com}
}

\begin{document}

\twocolumn[{%
\renewcommand\twocolumn[1][]{#1}%
\maketitle
\begin{center}
    \centering
    \captionsetup{type=figure}
    \includegraphics[width=0.95\textwidth]{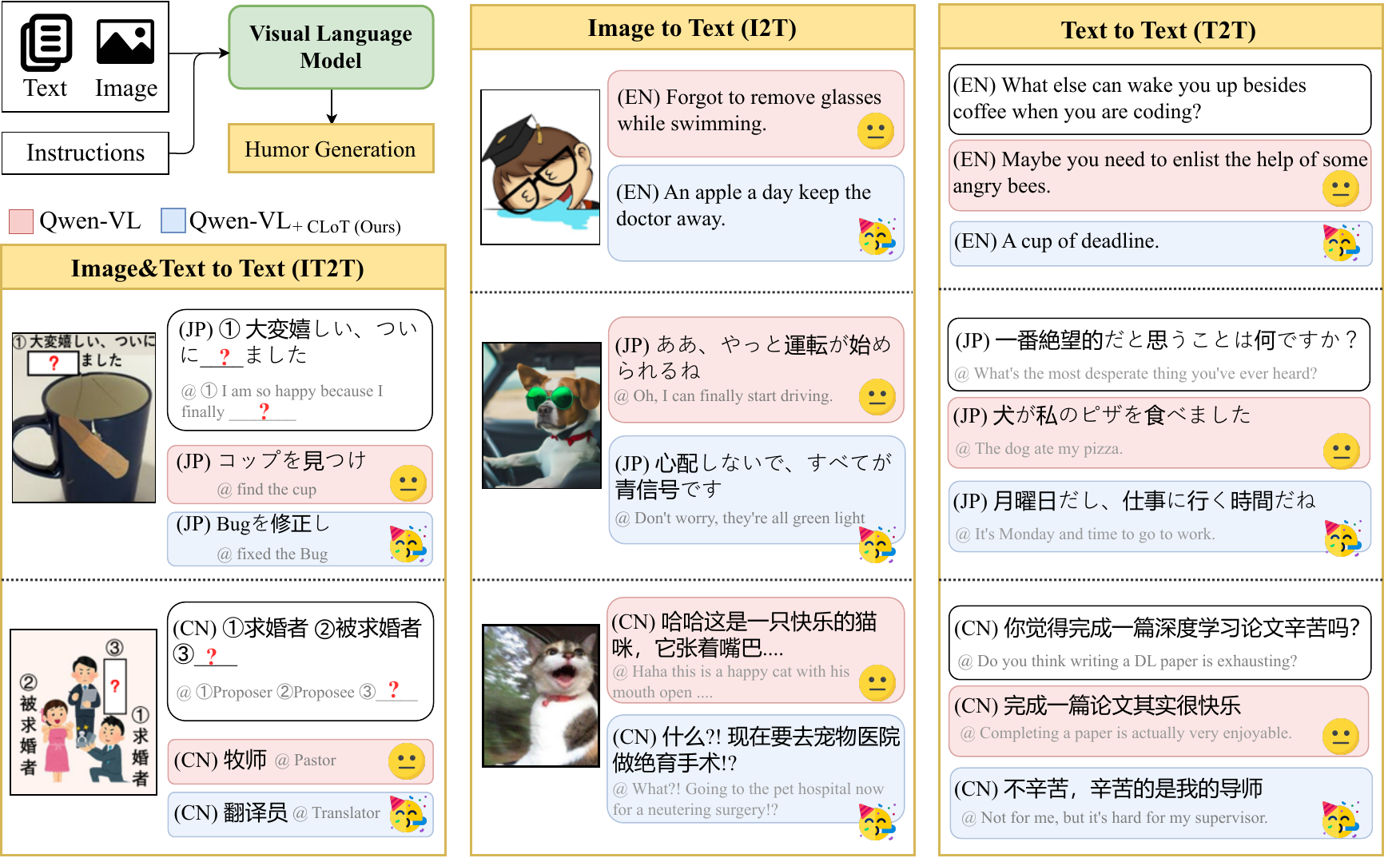}
    \captionof{figure}{
    Comparison between  (multimodal) large language model   (LLM, \textcolor{myblue}{\rule{0.25cm}{0.25cm}} red) and its CLoT-integrated version  ( \textcolor{myred}{\rule{0.25cm}{0.25cm} blue}) for Oogiri-style multimodal humor generation. According to the model input that can be image, text or both, there are three  Oogiri tasks,  ``Image\&Text to Text (IT2T)", ``Image to Text (I2T)", and ``Text to Text (T2T)”, where text can be English~(EN), Chinese~(CN), and Japanese~(JP). ``@" denotes translations in English. The baseline LLM  is Qwen-VL~\cite{Qwen-VL}.    While humor is subjective, these examples demonstrate CLoT's leap-of-thought capacity of using excellent creative thinking to produce high-quality humor responses. See more examples in Appendix. 
    } 
    \label{fig:example1}
\end{center}%
}]

\input{sec/0_abstract}

\input{sec/1_intro}

\input{sec/2_formatting}

{
    \small
    	\bibliographystyle{unsrt}
    \bibliography{main}
}
\input{sec/X_suppl}

\end{document}

%% file: sec/0_abstract.tex
\begin{abstract}
Chain-of-Thought (CoT)~\cite{kojima2022large,wei2022chain} guides large language models (LLMs) to reason step-by-step, and can motivate their logical reasoning ability. While effective for logical tasks, CoT is not conducive to creative problem-solving which often requires out-of-box thoughts and is crucial for innovation advancements. In this paper, we explore the Leap-of-Thought (LoT) abilities within LLMs — a non-sequential, creative paradigm involving strong associations and knowledge leaps. To this end, we study LLMs on the popular Oogiri game which needs participants to have good creativity and strong associative thinking for responding unexpectedly and humorously to the given image, text, or both, and thus is suitable for LoT study. Then to investigate LLMs' LoT ability in the Oogiri game,  we first build a multimodal and multilingual Oogiri-GO dataset which contains over 130,000 samples from the Oogiri game, and observe the insufficient LoT ability or failures of most existing LLMs on the Oogiri game.  Accordingly, we introduce a creative Leap-of-Thought (CLoT) paradigm to improve  LLM's LoT ability.  CLoT first formulates the Oogiri-GO dataset into LoT-oriented instruction tuning data to train pretrained LLM for achieving certain LoT humor generation and discrimination abilities.  Then CLoT designs an explorative self-refinement that encourages the LLM to generate more creative LoT data via exploring parallels between seemingly unrelated concepts and selects high-quality data to train itself for self-refinement. CLoT not only excels in humor generation in the Oogiri game as shown in Fig.~\ref{fig:example1} but also boosts creative abilities in various tasks like ``cloud guessing game" and ``divergent association task". These findings advance our understanding and offer a pathway to improve LLMs' creative capacities for innovative applications across domains. The dataset, code, and models will be released online. \url{https://zhongshsh.github.io/CLoT/}.

\end{abstract}

%% file: sec/1_intro.tex
\addtocontents{toc}{\protect\setcounter{tocdepth}{-1}}
\section{Introduction}
\label{sec:intro}

\begin{figure}[t]
  \centering
 \includegraphics[width=0.9\linewidth]{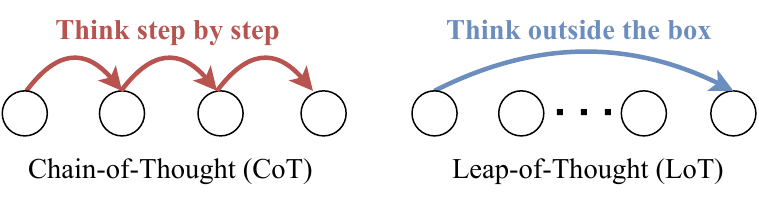}
  \caption{Comparison of  CoT and LoT.   ``$\bigcirc$" denotes the thought and ``\textrightarrow" represents  the connection between two thoughts. }
\label{fig:cotlot}
\vspace{-10pt}
\end{figure}

Large language models (LLMs)~\cite{touvron2023llama,bai2023qwen,vicuna2023,wei2021finetuned,saparov2022language,zeng2022socratic,driess2023palm,dong2023dreamllm,ye2023mplug,zhong2023adapter} have catalyzed a transformative era in problem-solving abilities, revolutionizing various domains within artificial intelligence. 
The advent of the Chain-of-Thought (CoT) paradigm~\cite{wei2022chain} and its further enhancements~\cite{zhang2022automatic,kojima2022large,yao2023tree,long2023large} have equipped these LLMs with a human-like step-by-step reasoning capacity. This augmentation has enabled LLMs to excel in intricate reasoning tasks spanning from language comprehension to visual understanding. As shown in Fig.~\ref{fig:cotlot} (Left), 
CoT instills LLMs with a sequential thinking process wherein each subsequent thought builds upon the previous one. 
This paradigm enhances the precision and rigor in logical processing, making it exceedingly effective for problems that demand closely linked logical reasoning.

However, the sequential nature of CoT might fall short in nurturing creativity and innovation, potentially limiting solutions in creative problem-solving scenarios~\cite{kahneman2011thinking,park2023papers}. 
For instance, proving an algebraic inequality often follows a step-by-step CoT process that progresses from one inequality to the next. Yet, an intuitive flash, e.g., a geometric interpretation, can yield a more creative solution. This type of insight, known as ``Leap-of-Thought" (LoT)~\cite{talmor2020leap,callaway2013cognitive}, a.k.a. mental leap~\cite{holyoak1995mental,olson1981leap,hofstadter1995review,holyoak1996mental}—the art of non-sequential thinking by association, drawing parallels between seemingly unrelated concepts, and facilitating a ``leap" of knowledge transfer.  
In contrast to CoT reasoning, LoT as depicted in Fig.~\ref{fig:cotlot} (Right), fosters associative reasoning and encourages thinking outside the box, which bridges disparate ideas and facilitates conceptual leaps. 
Embracing LLMs with a strong LoT ability can unlock significant potential for innovation, contributing to advancements in creative applications.
 
\begin{figure}[t]
  \centering
 \includegraphics[width=0.9\linewidth]{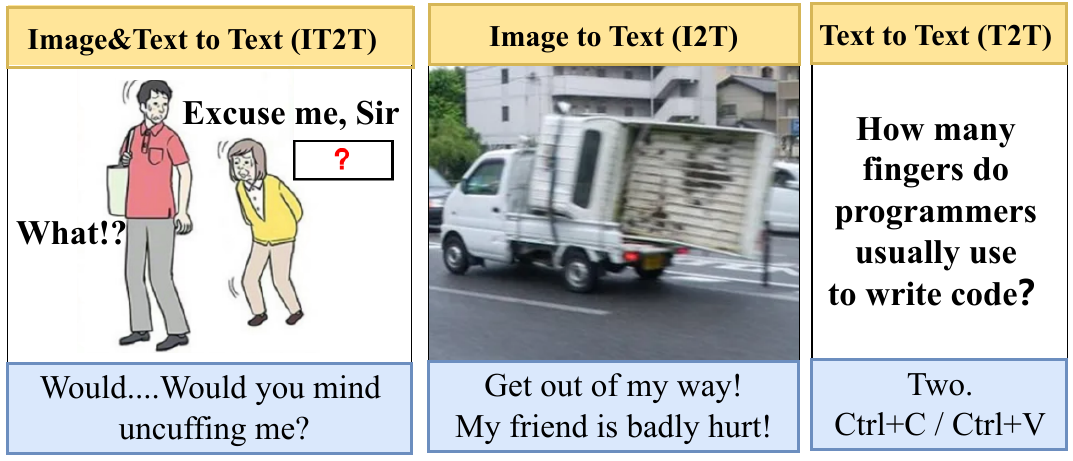}
  \caption{Examples of the three types of LoT-based Oogiri games. Players are required to make surprising and creative humorous responses (blue box) to the given multimodal information e.g., images, text, or both.}
\label{fig:oogiri}
\vspace{-10pt}
\end{figure}

In this paper, we aim to initially explore and enhance the LoT ability of LLMs.
However, thoroughly assessing LoT is challenging due to the complexity of measuring creative thinking~\cite{kitto1994measuring,molle1999eeg,jiang2014development} and the difficulty in gathering pertinent data, since generating novel ideas is challenging, even for humans~\cite{kahneman2011thinking}.
Given these constraints, we propose studying LoT in LLMs through the lens of Oogiri-style humor generation. Oogiri, a traditional Japanese creative game~\cite{oogiri}, requires participants to provide unexpected and humorous responses to prompts in the form of images, text, or a combination of both, as shown in Fig.~\ref{fig:oogiri}. 
This game challenges LLMs to demonstrate a sudden burst of insight and strong associative thinking, presenting a unique challenge for CoT-based methods, making it an ideal testbed for assessing the leap-of-thought abilities of LLMs.
Moreover, the extensive online presence of Oogiri guarantees a wealth of human-generated creative content, ideal for compiling an expansive leap-of-thought dataset.


To investigate the LoT ability of LLMs in the Oogiri game, we present the multilingual and multimodal Oogiri-GO dataset 
which comprises more than 130,000 high-quality Oogiri samples in English, Chinese, and Japanese, and curated to prompt textual humor in response to inputs that can be images, text, or both. 
Through extensive experiments, we discover that even the advanced LLMs and reasoning frameworks \cite{wang2023cogvlm,touvron2023llama,vicuna2023,kojima2022large}, such as GPT-4 and CoT, despite their exceptional reasoning capabilities, possessing a rich prior knowledge of diverse forms of humor \cite{kojima2022large}, still struggle to exhibit sufficient LoT ability for creative humor generation. Moreover, directly fine-tuning LLMs on the Oogiri-GO is not easy to improve the LoT ability. The more efficient utilization of humorous knowledge is needed to help LLM elicit creative responses.

Motivated by the human mental leap exercise process of ``remote association \& self-refinement"~\cite{Lee2012}, to enable LLMs with strong LoT ability for creation, we propose the Creative Leap-of-Thought (CLoT) paradigm which relies on two LoT-boosting stages. The first one is the associable instruction tuning stage which designs an associable instruction template to formulate the Oogiri-GO dataset into instruction data and trains an LLM to improve its LoT ability. The core here is the instruction template with a dual purpose: it randomly provides LLM with clues to establish connections between game inputs and creative responses, while also introducing empty clues to encourage LLM for unrestrained exploration and remote association thinking.

The second stage is explorative self-refinement which encourages the LLM to generate more creative LoT data via exploring parallels between seemingly unrelated concepts under weakly-associated conditions, and selects high-quality data to train itself for self-refinement.  These weakly-associated conditions can either be empty, or randomly sampled from an object noun set collected from the Oogiri-GO dataset. The former empty conditions to allow LLM to operate freely, and the latter ones help the LLM to link seemingly-unrelated and weakly-related concepts, and encourage the LLM to explore knowledge outside of traditional cognitive limitations.  This exploration strategy can help generate diverse high-quality data for self-refinement. 

Experimental results show that CLoT can greatly enhance the LoT ability of LLMs like Qwen~\cite{Qwen-VL}  and CogVLM~\cite{wang2023cogvlm}  across several types of Oogiri games. Specifically,  CLoT can help LLMs to generate much better humors in Fig.~\ref{fig:example1}. Moreover, CLoT-integrated LLMs achieve higher quantitative performance than the corresponding vanilla and CoT-integrated LLMs across the multiple-choice and ranking questions in the Oogiri game.  Also, CLoT can boost creative abilities on other tasks like  ``cloud guessing game" and ``divergent association task"~\cite{rombach2022high,zhang2023adding,olson2021naming}, showing its remarkable generalization ability.

\section{Related Works}
\label{sec:related}

\noindent\textbf{(1) Oogiri game} \begin{CJK*}{UTF8}{goth}(大喜利)\end{CJK*}  is a general term for a series of traditional Japanese comedy games.
In ancient times, there were different types of Oogiri, such as actors performing sumo wrestling, telling ghost stories, etc. The modern Oogiri game mainly refers to one specific type known as  Tonchi \begin{CJK*}{UTF8}{goth}(頓智)\end{CJK*}, typically presented in the format of game shows or intellectual quiz programs~\cite{oogiri}. Players are provided with various multimodal contents, which can be simple questions, random images, etc., and are then prompted to come up with humorous, creative responses to achieve surprising comedic effects, as the examples are shown in Fig.~\ref{fig:oogiri}. 
It is worth noting that the character ``\begin{CJK*}{UTF8}{goth}頓\end{CJK*}" in both Japanese and Chinese denote ``sudden", while ``\begin{CJK*}{UTF8}{goth}智\end{CJK*}" means ``intelligence, insight or intuition". This highlights the connection between the Oogiri game and the requirement for strong associative abilities in LoT, making Oogiri an ideal platform for exploring LoT capabilities within LLMs.

\noindent\textbf{(2) Multimodal LLMs and their creativity.}  Recently, multimodal Language Models~\cite{liu2023improvedllava,chen2023minigptv2,wang2023cogvlm,Qwen-VL} have garnered significant attention, particularly due to their impressive reasoning abilities~\cite{wei2021finetuned,saparov2022language,zeng2022socratic,driess2023palm,dong2023dreamllm,ye2023mplug,xing2023toa}. Moreover, there is a growing focus on exploring the creativity~\cite{ling2023unleashing,summers2023brainstorm,sun2023inspire,bhavya2023cam} of LLMs for applications such as scientific discovery~\cite{park2023papers,kang2022augmenting,hope2022scaling,liang2021stiffness,huang2023fast}, creative writing~\cite{swanson2021story,chakrabarty2022help,wu2022promptchainer,mirowski2023co,dang2023choice}, etc.

\noindent\textbf{(3) Computational humor} is a branch of computational linguistics and artificial intelligence that uses computers in humor research~\cite{binsted2006computational}, which encompasses various tasks, including humor detection~\cite{shahaf2015inside,tanaka2022learning,xu2022hybrid,chen2022integrating,kumar2022deephumor,wu2021mumor,ofer2022cards,xie2023funqa}, humor interpretation~\cite{xie2023funqa,hwang2023memecap,evans2019gender,vasquez2021cats}, and humor generation~\cite{li2023oxfordtvg,amin2020survey,zhang2020let,hossain2020stimulating,valitutti2013let,chaudhary2021towards}, etc. With the advancement of generative LLMs~\cite{touvron2023llama,wang2023cogvlm,Qwen-VL}, humor generation has become a popular focus while humor generation still faces challenges such as insufficient punchlines~\cite{popova2023does} and limited in multimodal contexts~\cite{chauhan2023mhadig,pramanick2022multimodal}.

\noindent\textbf{(4) Chain-of-Thought based Methods} provide the models with ``chain of thoughts"~\cite{wei2022chain,zhang2022automatic,kojima2022large,yao2023tree,long2023large}, i.e., reasoning exemplars~\cite{wei2022chain}, or a simple prompt ``Let's think step by step"~\cite{kojima2022large}, to encourage LLMs to engage in reasoning rather than simply providing answers directly~\cite{huang2022towards}.

%% file: sec/2_formatting.tex
\begin{table}[htbp]
\vspace{-5pt}
  \centering
  \resizebox*{0.85\linewidth}{!}{
    \begin{tabular}{ccccc}
    \toprule
    \textbf{Category} & \textbf{English} & \textbf{Chinese} & \textbf{Japanese}  & \textbf{Total} \\
    \midrule
    I2T   & 17, 336 & 32, 130 & 40, 278  &89, 744 \\
    T2T   & 6, 433  & 15, 797 & 11, 842   &34, 072 \\
    IT2T  & ---     & 912   & 9, 420    &10, 332 \\
    \bottomrule
    \end{tabular}
   } 
        \caption{Data distribution of the Oogiri-GO dataset.  For the IT2T task, its English version is not available due to cultural preference.}
  \label{tab:oogiridata}
\end{table}
\vspace{-0.5cm}

\section{Oogiri-GO Dataset}
\label{sec:oogirigo}
As introduced in Sec.~\ref{sec:related},  in the Oogiri game,   the participants need to unexpectedly and humorously respond to the given images,  text, or both. See three types of examples in Fig.~\ref{fig:oogiri}.  This game requests a sudden burst of insight and strong associative thinking to the given context, and provides an ideal platform to assess the leap-of-thought (LoT) ability of LLMs.  Accordingly, we collect Oogiri game data to build a large-scale  Oogiri-GO dataset which serves as a  benchmark to evaluate and improve LoT ability.

Specifically, Oogiri-GO is a multimodal and multilingual humor dataset, and contains more than 130,000 Oogiri samples in English, Chinese, and Japanese.  Notably,  in Oogiri-GO, 77.95\% of samples are annotated with human preferences, namely the number of likes, indicating the popularity of a response.  
As illustrated in Fig.~\ref{fig:oogiri},  Oogiri-GO contains three types of Oogiri games according to the input that can be images, text, or both,  and are respectively called   ``Text to Text" (T2T), ``Image to Text"  (I2T),  and ``Image \& Text to Text " (IT2T) for brevity.  See more examples in  Fig.~\ref{fig:example1}.  Table \ref{tab:oogiridata} summarizes the distribution of these game types.  For training purposes, 95\% of the samples are randomly selected to construct the training dataset, while the remaining 5\% form the test dataset for validation and analysis.

To create the Oogiri-GO dataset, there are three main steps,  including online data collection, machine filtering by LLM, and manual screening.  Firstly, to collect sufficient data, we source Oogiri game data from the official  Oogiri game platform, Bokete, and other popular platforms, such as  Twitter and Weibo which also host some Oogiri-game-alike data. Then, to guard against the inclusion of bias, violence, explicit content, offensive language, etc., we have placed a strong emphasis on rigorous safety checks during both machine and manual screening.  We first use the multimodal LLM Qwen-VL \cite{bai2023qwen} to do the initial screening of the raw data by constructing safety-checking prompts. Then, manual checking is performed on the remaining data. See more details about the dataset creation in the Appendix.

 \begin{figure}[t]
  \centering
 \includegraphics[width=1.0\linewidth]{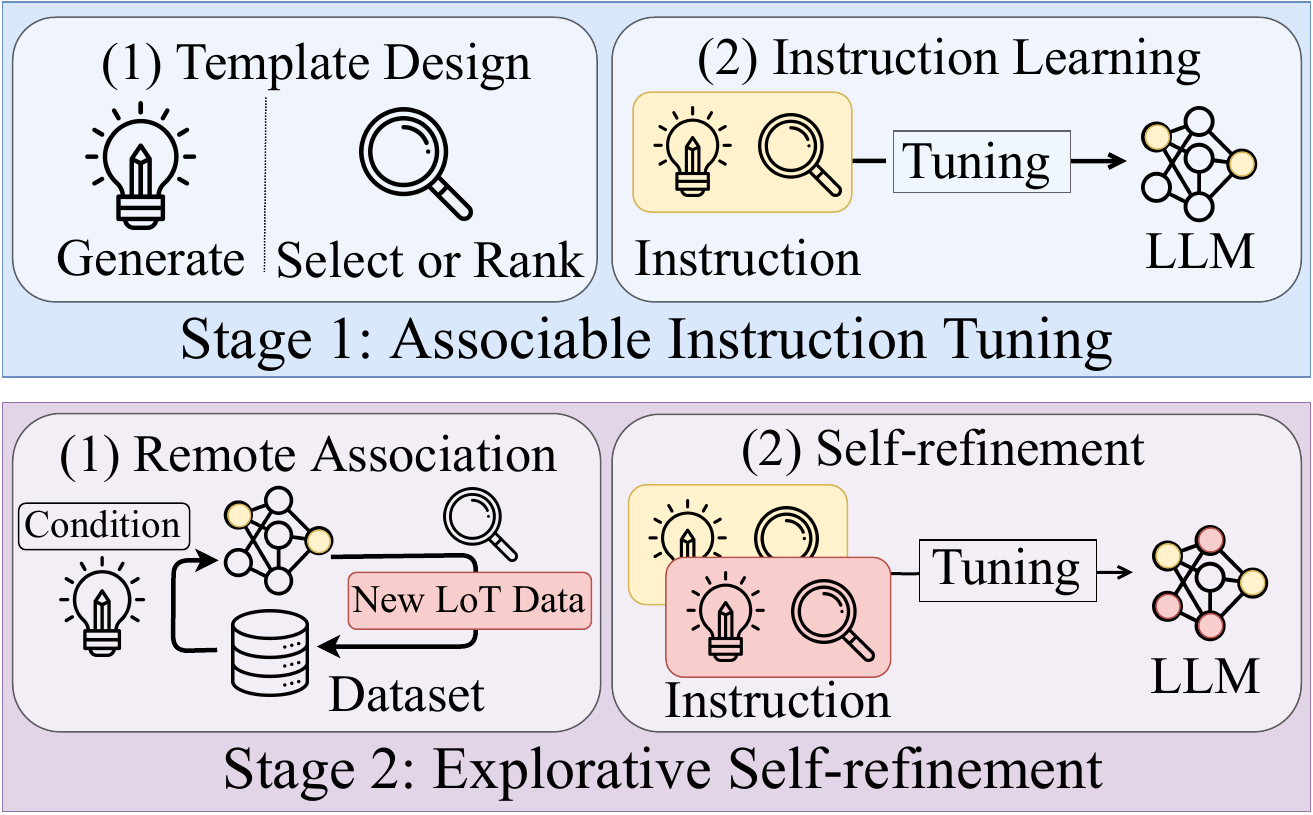}
  \caption{The overview of proposed Creative Leap-of-Thought.}
\label{fig:overview}
\vspace{-10pt}
\end{figure}

\section{Creative Leap-of-Thought (CLoT)}
\label{sec:method}
To augment the Leap-of-Thought (LoT) ability in (multimodal) Large Language Models (LLMs) for creative generation, we propose a novel Creative LoT framework (CLoT). As shown in Fig.~\ref{fig:overview}, CLoT relies on two  LoT-boosting stages. The first one is associable instruction tuning that formulates the Oogiri-GO dataset into instruction tuning data for training an LLM to improve its LoT ability (Sec.~\ref{sec:initial}).  The second one is explorative self-refinement that encourages the LLM to generate more creative LoT data via exploring parallels between seemingly unrelated concepts, and selects high-quality  data to train itself for self-refinement (Sec.~\ref{sec:remote}).  Finally, we present the CLoT inference to induce the LoT ability of the trained LLM (Sec.~\ref{CLoTsds}).

\subsection{Associable Instruction Tuning}
\label{sec:initial} 
LoT ability mainly includes associable generation and discrimination ability~\cite{Lee2012}. Given an input,  associable generation draws its parallels with seemingly unrelated concepts via remote association and then generates innovative responses, e.g., the unexpected humor for the Oogiri input.  Associable discrimination is to judge the matchiness among input and responses though they are seemingly unrelated, and then to select the most creative response.

Unfortunately, both associable generation and discrimination are not present in current LLMs, e.g., poor performance of  GPT4v~\cite{gpt4} in the Oogiri game observed in Sec.~\ref{sec:exp}. Moreover,  it is hard to improve these two LoT abilities via popular CoT-like prompt techniques. Indeed,  as shown in Sec.~\ref{sec:exp}, CoT even sometimes impairs the LoT performance of the LLMs like Qwen-VL~\cite{Qwen-VL} in the Oogiri game.

To address this issue, we propose associable instruction tuning which trains LoRA~\cite{hu2021lora} for LLMs on the Oogiri-GO dataset to achieve certain associable generation and discrimination abilities. It has two steps, including instruction generation and discrimination template design, and associable instruction learning.

\begin{figure}[h]
  \centering
 \includegraphics[width=0.75\linewidth]{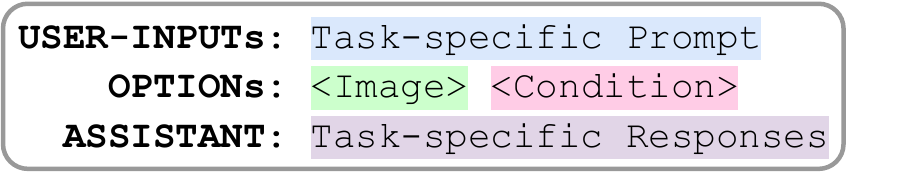}
    \vspace{-0.35cm}
  \caption{The LoT-oriented instruction templates.}
\label{fig:template}
\end{figure}

\noindent \textbf{(1) Instruction Generation \& Discrimination Templates}.  We  design LoT-oriented instruction templates to transform the Oogiri-GO dataset into instruction tuning data, and then train LLM to achieve associable generation and discrimination abilities.    
Our templates primarily comprise two components in Fig.~\ref{fig:template}: task-specific prompt and response.  For different abilities,  the templates need some special design.  

\vspace{-0.2cm}
\vspace{-0.3cm}

\begin{figure*}[t]
  \centering
 \includegraphics[width=1.00\linewidth]{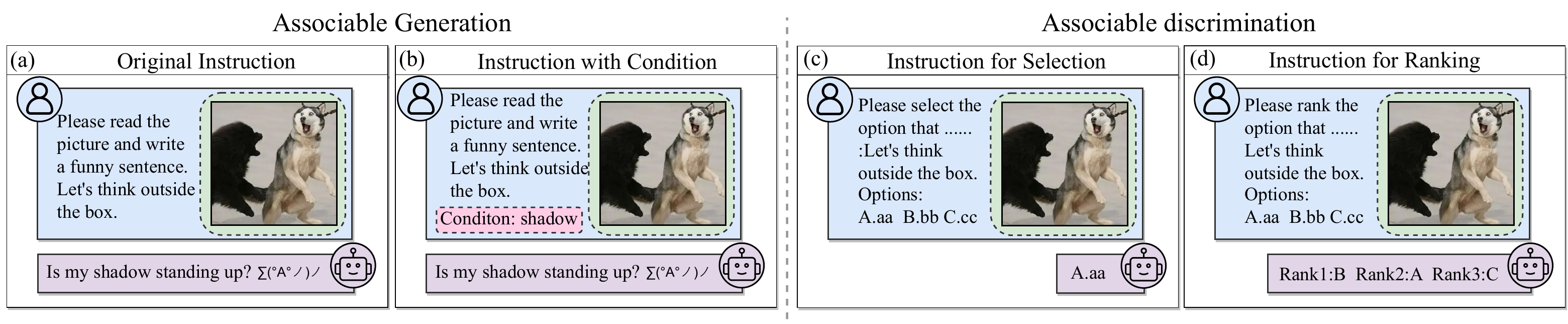}
  \caption{The details of LoT-oriented instructions templates. We take ``Image to Text" as an example, see the Appendix for the details of other categories' instructions. (a) and (b) are the instruction templates with/without conditions for associable generation. (c) and (d) are the two instructions about the selection and ranking of associable discrimination. All templates follow the formats in Fig.~\ref{fig:template}.}
\label{fig:format_all}
\end{figure*}

\underline{For associable generation},   ``USER-INPUTs" contains ``{Task-specific Prompt}" along with two optional   conditions,  ``Image" and ``Condition".  For ``Task-specific Prompt", we elaborately design several templates for different types of Oogiri game. See the Appendix for details and there is an image-2-text (I2T) Oogiri example in Fig.~\ref{fig:format_all}.   For ``Image" condition, it relies on the type of Oogiri game, e.g., being the image embeddings in I2T game and empty in T2T type. For the ``condition" option, it's set to empty with a probability of $\rho_c$,  and otherwise is randomly set as one noun in ``task-specific responses". This design gives the  LLM  a clue to connect the game input and the correct responses while also encouraging LLM to explore and unleash its creative thinking with probability $\rho_c$.  Finally,  ``{Task-specific Responses}"  are the ground truth responses of an Oogiri-GO data, and need to be predicted by LLM during training.  This task enforces the LLM to draw parallels between seemingly unrelated concepts in inputs and responses for giving innovative responses, e.g., the humor for the Oogiri input. This associable generation ability can assist the LLM to think outside the box and learn remote association thinking.

\underline{Regarding associable discrimination}, we aim to develop fundamental LoT discrimination skills for LLM. Based on the Oogiri-GO data, we design choice questions to enhance LLM's LoT discrimination ability, i.e., \textbf{selection} skill. Besides, as 77.95\% of the Oogiri-GO data have human preference annotations, i.e.,  the number of likes of several responses (see Sec.~\ref{sec:oogirigo}),   we design ranking questions to improve another discrimination skill, i,e., \textbf{ranking} ability.

For a choice question, as shown in Fig.~\ref{fig:format_all} (c), the options in ``Task-specific Prompt" contain the random permutations of ground truth response (GTR), image captions generated by BLIP2 \cite{li2023blip},  GTR from other images, rewrites of GTR by Qwen-14B \cite{bai2023qwen}.  See details in Appendix. For ``task-specific responses", it is the GTR. This design is to train LLM to improve its LoT selection ability.  For a ranking question, as shown in   Fig.~\ref{fig:format_all} (d), it is to enforce LLM  to rank multiple distinct responses of a given input to match their human preferences. By training on the choice and ranking questions, LLM is encouraged to distinguish LoT responses and align human creative preferences, improving its   LoT discriminative selection and ranking abilities.

 \noindent \textbf{(2) Associable Instruction Learning}.  By using the above instruction templates,  we augment the  130,000 samples in the Oogiri-GO dataset to more than 500,000 instructions whose formulation is in Fig.~\ref{fig:template}. During training,  LLM is required to predict the ``task-specific responses" according to the ``USER-INPUTs" which include ``Task-specific Prompt" and two additional optional conditions like image and text condition.  To avoid over-fitting, we only train standard LoRA~\cite{hu2021lora}  for the LLM with the associable instruction data. See more details in Appendix.

\subsection{Explorative Self-Refinement}
\label{sec:remote} 

After associable instruction tuning,  we aim to generate more high-quality creative data by LLM which are then used to train LLM for self-refinement.  To this end, we introduce an innovative stage called explorative self-refinement,   inspired by human LoT exercise process of ``remote association \& self-refinement", also known as mental leap~\cite{Lee2012,holyoak1995mental,holyoak1996mental}.  The remote association process refers to generating new ideas by associating remote concepts or thoughts, and self-refinement uses the generated data to enhance one's own LoT ability. In the following, we design two similar LoT exercise processes for LLM to improve its LoT ability.

\begin{table*}[htbp]
	
	\resizebox*{\linewidth}{!}{
    \begin{tabular}{lr|ccccc|ccccc|ccccc}
    \toprule
    \multirow{2}[4]{*}{Model} & \multirow{2}[4]{*}{Size} & \multicolumn{5}{c|}{Image\&Text to Text (IT2T)} & \multicolumn{5}{c|}{Image to Text (I2T)}    & \multicolumn{5}{c}{Text to Text (T2T)} \\
\cmidrule{3-17}          &       & 3T1   & 4T1   & 5T2   & Rank  & Avg.$\qquad$  & 3T1   & 4T1   & 5T2   & Rank  & Avg.$\qquad$  & 3T1   & 4T1   & 5T2   & Rank  & Avg.$\qquad$ \\
    \midrule
    GPT4v~\cite{gpt4} & -     & 19.3  & 14.9  & 3.2   & 56.7  & 23.5$\qquad$  & 29.1  & 15.1  & 3.9   & 60.4  & 27.1$\qquad$  & 27.1  & 16.8  & 6.8   & 53.5  & 26.1$\qquad$  \\
    LLaVA-1.5 \cite{liu2023improvedllava} & 13B   & 13.2  & 13.7  & 13.9  & 68.1  & 27.2$\qquad$  & 29.3  & 22.7  & 3.9   & 60.9  & 29.2$\qquad$  & 33.8  & 25.2  & 4.0   & 62.6  & 31.4$\qquad$  \\
    MiniGPT-v2~\cite{chen2023minigptv2} & 7B    & 6.1   & 3.4   & 4.0   & 60.7  & 18.6$\qquad$  & 5.3   & 4.0   & 3.8   & 60.5  & 18.4$\qquad$  & 10.8  & 7.3   & 3.5   & 59.4  & 20.3$\qquad$  \\
    mPLUG-Owl$_{\text{Multilingual}}$~\cite{ye2023mplug} & 7B    & 28.1  & 26.0  & 10.5  & 64.4  & 32.2$\qquad$  & 19.2  & 18.6  & 6.0   & 60.5  & 26.1$\qquad$  & 24.4  & 22.2  & 10.7  & 60.1  & 29.4$\qquad$  \\
    VisualGLM-6B~\cite{du2022glm} & 6B    & 24.1  & 22.5  & 9.7   & 67.4  & 30.9$\qquad$  & 14.3  & 20.4  & 8.8   & 61.9  & 26.4$\qquad$  & 13.1  & 20.2  & 7.1   & 61.3  & 25.4$\qquad$  \\
    \midrule
    Qwen-VL~\cite{Qwen-VL} & 7B    & 30.2  & 26.0  & 10.4  & 67.7  & 33.6$\qquad$  & 23.2  & 23.1  & 11.9  & 62.2  & 30.1$\qquad$  & 23.4  & 25.0  & 13.3  & 59.6  & 30.3$\qquad$  \\
    Qwen-VL$_{+\text{AIT (Ours)}}$ & 7B    & 39.7  & \textbf{38.9} & 15.7  & 67.3  & 40.4{$_{+\ 6.8}$}  & 38.8  & 30.5  & 15.7  & 62.3  & 36.8{$_{+\ 6.7}$}  & 30.6  & 28.7  & 16.7  & 62.6  & 34.6{$_{+\ 4.3}$}  \\
    Qwen-VL$_{+\text{CLoT (Ours)}}$ & 7B    & \textbf{41.8} & 38.7  & \textbf{21.6} & \textbf{68.5} & \textbf{42.7}{\color{red}$_{+\ 9.1}$} & \textbf{39.8} & \textbf{35.1} & \textbf{22.7} & \textbf{64.4} & \textbf{40.5}{\color{red}$_{+10.4}$} & \textbf{38.8} & \textbf{29.4} & \textbf{21.0} & \textbf{64.7} & \textbf{38.5}{\color{red}$_{+\ 8.2}$} \\
    \bottomrule
    \end{tabular}%
	}
 \centering
	\caption{The accuracy (\%) of choice questions and the NDCG (\%) of ranking questions on  \textbf{mutilmodal multilingual models}. $m$T$n$ choice question selects $n$ correct answers from  $m$  options. ``Avg.'' is the average of all metrics. ``AIT" denotes associable instruction tuning. 
 	\vspace{-1em}
}
	\label{tab:vlm}%
\end{table*}%

 \noindent \textbf{(1) Explorative Remote Association}. The core here is to prompt the LLM to generate a diverse array of creative responses under weakly-associated conditions. To implement this, we extract a set of object nouns, denoted as $\Sm$, from the text in the Oogiri-GO training data. See details in Appendix. Then, for each user-input $\Imm$ (see Fig.~\ref{fig:template}),   
 we generate $n$ weakly-associated conditions $\{\Cm_i\}_{i=1}^n$. These conditions can either be empty with a probability $\rho\in(0,1)$ to give freedom to LLM, or uniformly randomly sampled from the noun set $\Sm$ to enforce LLM to build connections between different concepts.  
 Next, we add the condition $\Cm_i$ into user-input $\Imm$,  and feed $\Imm$ into the LLM to generate a humor candidate $\Rm_i$. Repeating this process with different conditions $\Cm_i$ can generate  a total of $n$ candidates  $\{\Rm_i\}_{i=1}^n$.

 Then the LLM  ranks these candidates by its discriminative ranking ability learned in Sec.~\ref{sec:initial}. Next, it mixes the top-2 candidates with the ground truth responses (GTR), and selects the top-1 as the final response. Finally, if the selected top-1 response is the GTR,   we discard this sample. Here first filtering out low-quality responses can improve the accuracy of subsequent top-1 selection, since $(n+1)$-choice problem is often more challenging than $3$-choice problem as shown in Sec.~\ref{sec:exp}.  
 By repeating this process, we progressively gather sufficient new high-quality data.

 The core of this approach is the weakly-associated conditions  $\{\Cm_i\}_{i=1}^n$ which can encourage the LLM to engage in remote associations. This is because the empty conditions allow LLM to operate freely, while the object noun conditions compel the LLM to draw connections between seemingly unrelated concepts. This mechanism facilitates the establishment of links between seemingly-unrelated and weakly-related concepts, encouraging the LLM to explore knowledge outside of traditional cognitive limitations. The exploration ability distinguishes our CLoT from CoT which primarily guides the LLM to exploit its inherent reasoning ability without emphasizing knowledge exploration.

  \noindent \textbf{(2) Self-refinement}. Here we combine the above generated instructions with vanilla instruction tuning samples in Sec.~\ref{sec:initial} to form a dataset with more than 550,000 samples to train our LLM again. Since the above generated data is of high diversity because of its exploration strategy,  they prevent performance collapse~\cite{hataya2023contamination,shumailov2023model} during self-refinement phase, and can improve the LoT performance across several creative tasks as shown in Sec.~\ref{sec:exp}. See the ablation study and more discussions in Sec.~\ref{sec:ablation}.

\begin{algorithm}[t]
    \caption{Inference Step of CLoT}
    \scalebox{0.9}{%
        \begin{minipage}{\linewidth}
            \textbf{Input:} Input $\Imm$, CLoT-trained LLM $\mathcal{A}$, response number $n$
    
            \textbf{Output:} Creative response $\Rm$.
        
            \begin{algorithmic}[1]
                \State\algorithmiccomment{Creating the candidate responses }
                \State construct $n$ weakly-associated conditions $\{\Cm_i\}_{i=1}^n$
                \State $\{\Rm_i\}_{i=1}^n \gets\mathcal{A}([\Imm, \{\Cm_i\}_{i=1}^n ])$
                
                \State\algorithmiccomment{Choosing most creative response}
                \State Top-2 $\Rm^\prime_1, \Rm^\prime_2 \gets \mathcal{A}([\Imm, \{\Rm_i\}_{i=1}^n])$ with ranking ability
                \State Best $\Rm \gets \mathcal{A}([\Imm, \Rm^\prime_1, \Rm^\prime_2])$ with selection ability
                
                \State\Return Best response $\Rm$.
            \end{algorithmic}
        \end{minipage}%
    }
    \label{alg:clot}
\end{algorithm}

\subsection{CLoT Inference}\label{CLoTsds} 

After the two LoT-boosting phases in Sec.~\ref{sec:initial} and \ref{sec:remote}, the LLM acquires sufficient LoT ability. Now we introduce the inference steps of LLM to release its LoT ability. 
Formally, given an Oogiri  user-input $\Imm$ of the formation in Fig.~\ref{fig:template}, LLM first uses explorative remote association  in Sec.~\ref{sec:remote} to construct $n$ weakly-associated conditions, and then  follows  Sec.~\ref{sec:remote} to generate $n$ responses $\{\Rm_i\}_{i=1}^n$. 
 Next, LLM ranks these responses by using its learned ranking skill and finally selects the top-1 one from the ranked top-2 response via its selection skill.   The reason to first use ranking before selection is that as shown by experimental results in Sec.~\ref{sec:exp},  directly choosing the best one from a large number of options has poor accuracy, and ranking can filter out low-quality candidates to improve the selection accuracy.  See Algorithm~\ref{alg:clot} for an overview of CLoT inference steps.

\begin{table*}[htbp]
  \centering
	
\vspace{-1em}
    \resizebox*{1.00\linewidth}{!}{
    \setlength{\tabcolsep}{3mm} 
    \begin{tabular}{lr|cccccc|cccccc}
    \toprule
    \multirow{2}[4]{*}{Model} & \multirow{2}[4]{*}{Size} & \multicolumn{6}{c|}{Image to Text (I2T)}            & \multicolumn{6}{c}{Text to Text (T2T)} \\
\cmidrule{3-14}          &       & 2T1   & 3T1   & 4T1   & 5T2   & Rank  & Avg.$\qquad$  & 2T1   & 3T1   & 4T1   & 5T2   & Rank  & Avg.$\qquad$ \\
    \toprule
    InstructionBLIP~\cite{InstructBLIP} & 13B   & 19.8  & 13.7  & 15.5  & 1.1   & 65.5  & 23.1$\qquad$  & 22.3  & 16.0  & 17.0  & 0.7   & 59.5  & 23.1$\qquad$ \\
    mPLUG-Owl$_{\text{LLaMA2}}$~\cite{ye2023mplug} & 7B    & 22.3  & 12.7  & 15.0  & 4.2   & 59.9  & 22.8$\qquad$  & 24.2  & 13.7  & 12.6  & 3.1   & 59.2  & 22.6$\qquad$ \\
    Otter~\cite{li2023otter} & 7B    & 15.8  & 9.9   & 8.5   & 7.1   & 61.3  & 20.5$\qquad$  & 3.8   & 3.3   & 4.8   & 5.4   & 58.5  & 15.1$\qquad$ \\
    \midrule
    CogVLM-17B~\cite{wang2023cogvlm} & 7B    & 37.6  & 26.4  & 18.3  & 2.5   & 64.6  & 29.9$\qquad$  & 35.1  & 27.8  & 24.8  & 7.5   & 64.1  & 31.9$\qquad$ \\
    CogVLM-17B$_{+\text{AIT (Ours)}}$ & 7B    & 57.4  & 37.4  & 33.5  & 21.8  & 64.6  & 42.9$_{+13.1}$  & 55.4  & 46.5  & 26.4  & 18.2  & 64.4  & 42.2$_{+10.3}$ \\
    CogVLM-17B$_{+\text{CLoT (Ours)}}$ & 7B    & \textbf{66.9} & \textbf{47.6} & \textbf{43.4} & \textbf{30.7} & \textbf{69.4} & \textbf{51.6}{\color{red}$_{+21.7}$} & \textbf{64.8} & \textbf{52.9} & \textbf{33.6} & \textbf{21.8} & \textbf{68.6} & \textbf{48.3}{\color{red}$_{+16.4}$} \\
    \bottomrule
    \end{tabular}%
   }
   \caption{The accuracy (\%) of choice questions and the NDCG (\%) of ranking questions on various \textbf{mutilmodal non-multilingual models} (English). See notations in Table \ref{tab:vlm}. We only consider I2T and T2T since English IT2T is not available due to cultural preference. }
  \label{tab:non-multilingual}%
  \vspace{-10pt}
\end{table*}%

\begin{table}[t]
  \centering
  
    \resizebox*{1.00\linewidth}{!}{
    \begin{tabular}{lr|ccccc}
    \toprule
    Model  & Size & 3T1   & 4T1   & 5T2   & Rank  & Avg. \\
    \midrule
    GPT-3.5~\cite{gpt4} & -     & 45.3  & 30.4  & 6.7   & 61.6  & 36.0 \\
    GPT-4~\cite{gpt4} & -     & 49.2  & 20.4  & 3.6   & 54.7  & 32.0 \\
    \midrule
    \multirow{3}[2]{*}{LLAMA2~\cite{touvron2023llama}} & 7B    & 18.9  & 13.5  & 1.1   & 60.4  & 23.5 \\
          & 13B   & 15.6  & 20.0  & 1.8   & 60.5  & 24.5 \\
          & 70B   & 27.8  & 16.1  & 3.8   & 62.0  & 27.4 \\
    \midrule
    \multirow{2}[2]{*}{Baichuan2~\cite{baichuan2023baichuan2}} & 7B    & 28.3  & 22.6  & 11.6  & 64.6  & 31.8 \\
          & 13B   & 21.7  & 18.3  & 8.9   & 61.5  & 27.6 \\
    \midrule
    \multirow{2}[2]{*}{Qwen~\cite{bai2023qwen}} & 7B    & 23.1  & 20.4  & 8.0   & 61.4  & 28.2 \\
          & 14B   & 27.4  & 22.2  & 12.3  & 59.5  & 30.3 \\
    \midrule
    ChatGLM3~\cite{du2022glm} & 6B    & 15.6  & 17.0  & 5.4   & 59.4  & 24.3 \\
    \midrule
    \multirow{2}[2]{*}{Vicuna-v1.5~\cite{vicuna2023}} & 7B    & 32.6  & 23.5  & 0.0   & 63.0  & 29.8 \\
          & 13B   & 30.2  & 23.0  & 2.7   & 62.2  & 29.5 \\
    \midrule
    Qwen-VL$_{+\text{CLoT (Ours)}}$ & 7B    & 51.7  & 32.3  & \textbf{24.8} & 65.0  & 43.4 \\
    CogVLM-17B$_{+\text{CLoT (Ours)}}$ & 7B    & \textbf{52.9} & \textbf{33.6} & 21.8  & \textbf{68.6} & \textbf{44.2} \\
    \bottomrule
    \end{tabular}%
   }
   \caption{The accuracy (\%) of choice questions and the NDCG (\%) of ranking questions on various \textbf{large language models}. Here we use English T2T task for test.   See notations in Table \ref{tab:vlm}. 
  	\vspace{-1em}
 }
  \label{tab:llm}%
  \vspace{-10pt}
\end{table}%

\section{Experiments}
\label{sec:exp}

\subsection{Evaluation Questions and Metrics}
Inspired by the humor benchmarks in \cite{hessel2023androids}, we first develop choice and ranking questions as introduced in Sec.~\ref{sec:initial} (see examples in Fig.~\ref{fig:format_all} (c-d)),  and then quantitatively evaluate the LoT ability of LLMs  on the  Oogiri-GO  test dataset.  For the \textit{choice questions}, $m$T$n$ for short, they need LLMs to choose $n$ ``leap-of-thought'' humor responses from $m$  options given the input.  Here we build four types of $m$T$n$ questions, including 2T1, 3T1, 4T1, and 5T2. 2T1 means two options, the ground-truth response (GTR) and an image caption generated by BLIP2 \cite{li2023blip}. 3T1 adds unrelated answers, e.g., other image captions. 4T1 further adds the GTR rewrite by Qwen-14B~\cite{bai2023qwen}. 5T2 has an extra GTR. For these questions, their difficulty increases progressively, and is diverse to ensure comprehensive evaluation. For choice questions, we use accuracy as the evaluation metric. 
Additionally, for the questions in test set whose responses have ground-truth human preference, e.g., the number of likes, we develop the \textit{ranking questions} that always rank five candidates.  For evaluation, we adopt the top-1 accuracy and the widely used ranking metric,i.e., Normalized Discounted Cumulative Gain (NDCG) \cite{jarvelin2002cumulated,radlinski2010comparing}. We provide more experimental details in the Appendix.

\subsection{Evaluation by Choice and Ranking Questions}

\noindent\textbf{Evaluation on Multimodal Multilingual LLMs.} We plug our associable instruction tuning (AIT) and our CLoT into the SoTA open-source multimodal multilingual model Qwen-VL~\cite{Qwen-VL} to obtain Qwen-VL$_{+\text{AIT}}$ and Qwen-VL$_{+\text{CLoT}}$, respectively.    
Table~\ref{tab:vlm} shows that,  on three tasks (IT2T, I2T and T2T)  which include English, Chinese and Japanese questions,  Qwen-VL achieves the best LoT performance among all baselines in most cases. In comparison, Qwen-VL$_{+\text{AIT}}$ achieves a noticeable improvement on the SoTA Qwen with average accuracy enhancements of 6.8\%, 6.7\%, and 4.3\% on the three tasks, respectively.
Importantly, Qwen-VL$_{+\text{CLoT}}$ further enhances Qwen-VL, showing improvements of 9.1\%, 10.4\%, and 8.2\% in accuracy across these tasks.
These results demonstrate the efficacy of the two stages in CLoT, i.e.,  associable instruction tuning and explorative self-refinement.

\noindent\textbf{Evaluation on Multimodal Non-multilingual LLMs.} Here we integrate our CLoT with the  SoTA multimodal non-multilingual model, CogVLM-17B~\cite{wang2023cogvlm}, and evaluate  it on the English I2T and T2T tasks. 
Table \ref{tab:non-multilingual} shows that CogVLM-17B$_{+\text{AIT}}$ achieves remarkable improvements over the standard CogVLM-17B, and CogVLM-17B$_{+\text{CLoT}}$ consistently demonstrates significantly superior performance compared to CogVLM-17B.

\noindent\textbf{Evaluation on Single-Modal LLMs.}   Now we test LLMs that can handle only pure texts, using the English T2T task for evaluation.  
Table~\ref{tab:llm} also indicates the insufficient   LoT ability within existing LLMs, ranging from small to large models.  
Fortunately, our CLoT significantly improves the LoT ability of these LLMs,
as demonstrated by the notable improvement in accuracy.

\begin{figure*}[h]
  \centering
 \includegraphics[width=1.00\linewidth]{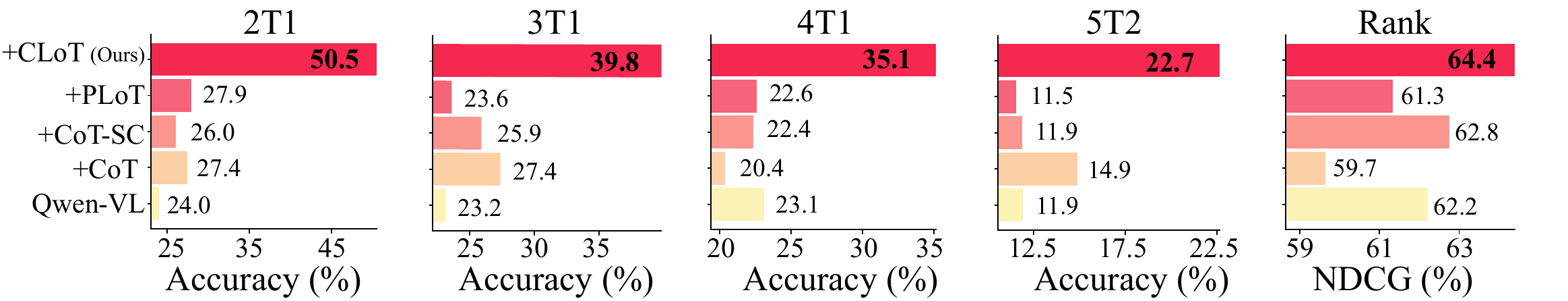}
 \vspace{-20pt}
  \caption{
  	The accuracy (\%) of choice questions and the NDCG (\%) of ranking questions on our CLoT and various \textbf{reasoning frameworks}. The baseline is Qwen-VL on multilingual I2T task.  For $m$T$n$ choice questions,  one needs to select $n$ correct answers from   $m$  options.
  	}
\label{fig:reasoning}
\vspace{-10pt}
\end{figure*}

\noindent\textbf{Comparison with CoT-alike Reasoning Frameworks.} We also find that existing reasoning frameworks are not as effective as CLoT in enhancing LoT ability.  Fig. \ref{fig:reasoning} compares CLoT with CoT~\cite{kojima2022large,wei2022chain}, CoT-SC~\cite{wang2022self}, and prompted-based LoT (PLoT) with the prompt ``let's think outside the box". The results reveal that CoT-alike frameworks do not enhance LoT performance of LLMs, while CLoT demonstrates the ability to consistently enhance LLMs. 

Our experiments and analysis reveal that, unlike CoT-based methods, LoT cannot be directly achieved by prompting alone. 
This is because the inherent reasoning capabilities and extensive knowledge of LLMs are not sufficient to enable LoT ability.
However, when trained with our proposed CLoT method, LLMs can effectively engage in a range of creative tasks. Additionally, the use of specific prompting techniques can enhance the LoT ability of CLoT-trained LLMs. These findings suggest that LoT could potentially be considered an additional general reasoning ability for LLMs
that is not contained in current LLMs.

\begin{figure}[t]
  \centering
 \includegraphics[width=0.99\linewidth]{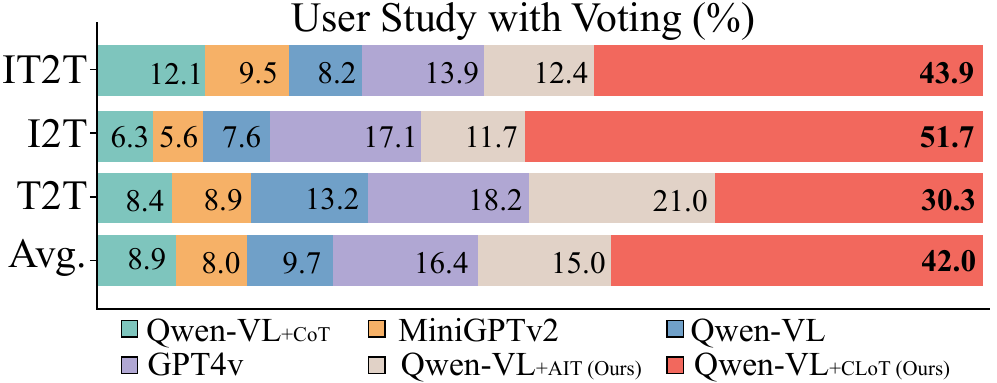}
  \caption{User study with voting (\%) for Oogiri-style creative responses by different models and improved methods.}
\label{fig:user-study}
\vspace{-10pt}
\end{figure}

\subsection{Human Evaluation}
We conduct a user preference study to test creativity of LLMs. Here we select six LLMs to generate responses for a total of eighteen questions across three tasks (IT2T, I2T and T2T). We use choice questions, and ask users to choose the most creative and humorous responses.     Fig.~\ref{fig:user-study} summarizes the statistical analysis of 154 valid surveys. The results show that users have a strong inclination towards selecting the results of CLoT across three tasks, highlighting the high-quality creative content generated by CLoT. See more user study details in Appendix.

\subsection{Evaluation on  Other Creative Tasks}

\begin{figure}[t]
  \centering
 \includegraphics[width=0.95\linewidth]{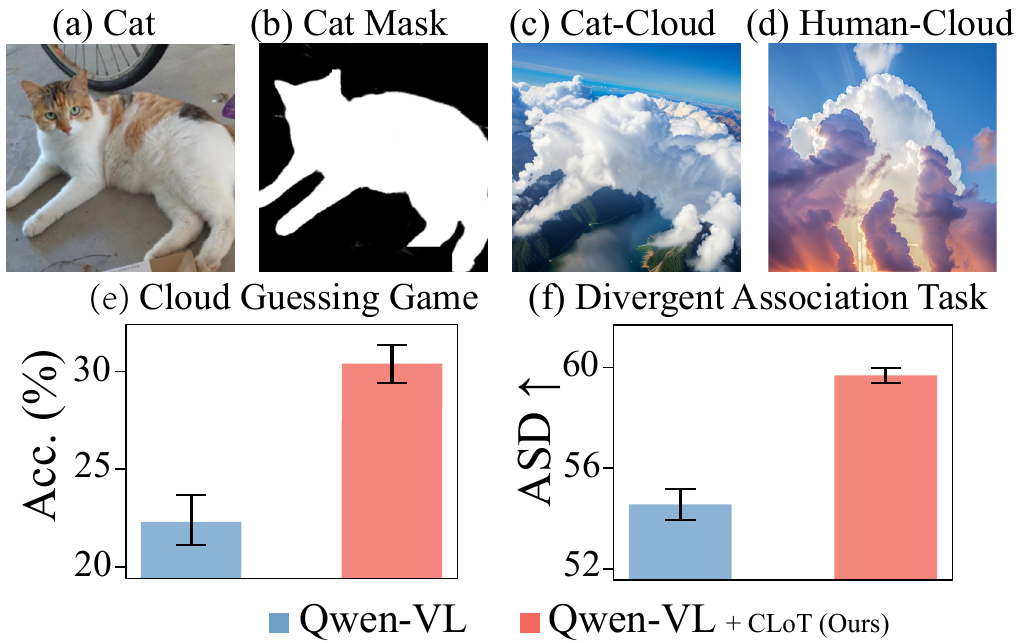}
 \vspace{-5pt}
  \caption{Evaluation of  CLoT on the creative CGG (e) and DAT (f) tasks.
  (c-d): examples of cloud guessing games. (b): conditional masks of image (a) for generating cloud images.
  }
\label{fig:generalization}
\vspace{-10pt}
\end{figure}

 To evaluate the generalization ability of CLoT, we test CLoT on another two creative tasks, including Cloud Guessing Game (CGG) and  Divergent Association Task (DAT).  In CGG, the LLM is to identify the shape of white clouds, and then to select the corresponding shapes from given options. For instance, the white clouds in  Fig.~\ref{fig:generalization} (c) has a shape of a cat, and the one in Fig.~\ref{fig:generalization} (d) is similar to a human. These white cloud images are generated by a control diffusion model \cite{rombach2022high,zhang2023adding,huang2023scalelong,shi2023exploring}, guided by masks shown in Fig.~\ref{fig:generalization} (b). We use top-1 accuracy as metric. See more details in Appendix.  
For DAT, it is a classic creativity test~\cite{beketayev2016scoring,olson2021naming} which needs participants to choose words with larger semantic distances
among 10 unrelated nouns.  Here for test easily, we transfer the DAT benchmark~\cite{olson2021naming} to a series of choice questions and take the standard average semantic distance~(ASD) as a metric. These questions can challenge the LLM to select the one word from nine options that differs from the given word most.  See more details in Appendix.  
CGG and DAT can test the LoT ability of LLMs, specifically their remote association thinking ability, and provide quite different evaluation platforms.  As shown in Fig.~\ref{fig:generalization} (e-f),  CLoT can also significantly enhance the performance of the SoTA Qwen-VL on both CGG and DAT tasks. Specifically, CLoT-integrated Qwen-VL improves the vanilla Qwen-VL by about 8\%  on the CGG task and 5\% on the DAT task. These results well demonstrate the good generalization and transferability of CLoT.

\subsection{Ablation Study}
\label{sec:ablation}
\noindent\textbf{Weakly-associated Conditions.} 
By default, to encourage  remote association, we use weakly-associated conditions randomly sampled from the noun set on the whole dataset in Sec.~\ref{sec:remote}.

\begin{figure}[t]
  \centering
 \includegraphics[width=0.9\linewidth]{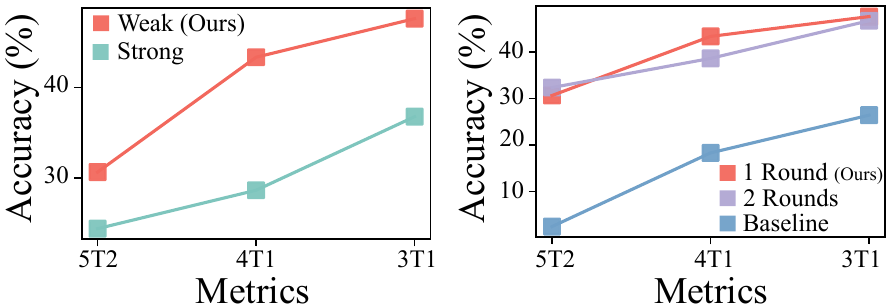}
 \vspace{-5pt}
  \caption{The ablation study of CLoT. We use  CogVLM as baseline on the English I2T task. \textbf{Left:} weakly-associated condition v.s. strongly-associated condition during explorative remote association. \textbf{Right:} The effect of rounds of self-refinement.}
\label{fig:self-refinement}
\vspace{-10pt}
\end{figure}

To verify the effectiveness of weakly-associated conditions,
now we resort to strongly-associated conditions sampled from the noun set of the current image caption.
Results in Fig.~\ref{fig:self-refinement} (Left) show that using weakly-associated conditions is superior and more conducive to fostering the creativity of LLMs.
The weakly-associated conditions enable the LLM
to generate more diverse LoT responses, while the strong clue from the strongly-associated conditions limit the diversity of LoT generations.

\noindent\textbf{Round of Self-refinement.} 
By default, we run one-round self-refinement for the Oogiri game. Here we explore whether more rounds of the self-refinement can further improve the LoT ability. 
Fig.~\ref{fig:self-refinement} (Right) shows that a single round of self-refinement already yields promising performance, whereas additional rounds do not yield significant further improvements.
As shown in Fig.~\ref{fig:self-refinement} (Left), the diversity of the condition set $\Sm$ is crucial to self-refinement, since it decides whether the associable remote stage can generate high quality and diverse data.  However, the condition set is not expanded during the second-round self-refinement, which consequently limits further improvements in performance.  Effectively increasing the scale of the condition set is an effective way for further improvement. See more discussion in Appendix. But its exploration falls outside the scope of this work and is left for our future research.

\section{Conclusion}
\label{sec:ana}
In this paper,  we propose a Creative Leap-of-Thought (CLoT) paradigm to improve  LLM's leap-of-thought (LoT) ability.  CLoT first collects a multimodal  Oogiri-GO dataset, and formulates it into instruction tuning data to train LLM to improve its LoT ability.  Then CLoT designs an explorative self-refinement that lets LLM generate more creative LoT data via exploring parallels among different concepts and selects high-quality data to train itself for self-refinement.  Experimental results show the effectiveness and generalization ability of CLoT across several creative tasks.

%% file: sec/X_suppl.tex
\clearpage

\setcounter{page}{1}

\onecolumn
\begin{figure}[h]
  \centering
 \includegraphics[width=0.99\linewidth]{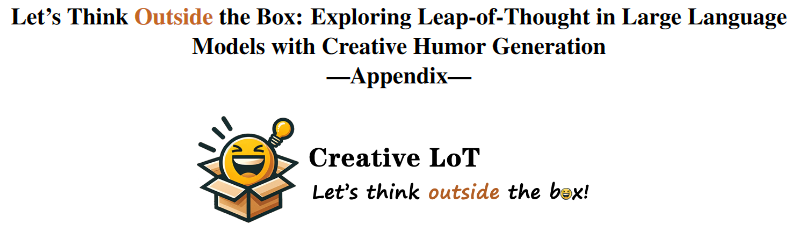}
\label{fig:overview22}
\vspace{-20pt}
\end{figure}
\renewcommand{\contentsname}{Contents}
\tableofcontents 
\addtocontents{toc}{\protect\hypersetup{linkcolor=black}}
\appendix

\addtocontents{toc}{\protect\setcounter{tocdepth}{3}}

\clearpage
\section{Introduction of Appendix}

 The appendix is structured as follows. 
 In Appendix \ref{sec:highlight}, we initially provide a detailed summary of the novelty in our paper and emphasize that our proposed CLoT is not tailored for humor generation. Instead, it focuses on the Leaf-of-Thought capability of large language models.
 In Appendix \ref{sec:moreexpdetail}, we further demonstrate the superiority of CLoT through various aspects. 
 Firstly, we showcase the performance of CLoT in each language, i.e., English, Chinese, and Japanese respectively, emphasizing its versatility across languages. 
Following that, to illustrate CLoT's impact on enhancing creativity, we present its ability to generate diverse creative solutions for the same Oogiri game data sample. 
Lastly, we provide additional generated humor responses of various types of Oogiri games for different LLMs. Appendix \ref{sec:aboutdata} outlines the construction details of the Oogiri-GO dataset, encompassing the data collection process and both machine and human-driven filtering processes. In Appendix \ref{sec:expdetails}, we meticulously detail main experiments presented in this paper, while Appendix \ref{sec:othercreative} provides a comprehensive description of other creative tasks. Furthermore, Appendix \ref{sec:selfrefinement} discusses two pivotal issues during the self-refinement phase, namely the round of refinement and potential performance collapse. Finally, in Appendix \ref{sec:further}, we address noteworthy concerns about the leaf-of-thought through a series of discussions.

\section{Highlight Perspective}
\label{sec:highlight}
\subsection{The \textcolor{logo2}{Novelty} of Our Paper.}

The novelty of this paper can be summarized as follows:
\begin{itemize}
  \item (\underline{\textbf{Pioneering}}) To the best of our knowledge, we are the first to profoundly explore the Leap-of-Thought (LoT) capability in multimodal large language models (LLMs). This involves challenging LLMs to think outside the box, a non-sequential thinking skill equally crucial alongside popular sequential thinking abilities, such as Chain-of-Thought based methods~\cite{wei2022chain,zhang2022automatic,kojima2022large,yao2023tree,long2023large}. The LoT ability serves as a cornerstone for creative exploration and discovery in LLMs.
  \item (\underline{\textbf{Large-scale Creative Dataset}}) Given the scarcity of large-scale datasets for investigating creativity in the current community and the inherent challenges in collecting creative data (refer to Appendix \ref{sec:oneround}), in this paper, we identify the Oogiri game as an ideal platform for exploring the LoT ability of LLMs (refer to Appendix \ref{sec:tailor}), and gather over 130,000 creative data samples about Oogiri game, forming a large-scale creative dataset named Oogiri-GO.
  \item (\underline{\textbf{Novel Paradigm for Improving LoT}}) Our experiments and analysis reveal that existing LLMs struggle to evoke LoT ability solely relying on their intrinsic reasoning abilities and extensive prior knowledge. Therefore, we propose a novel Creative Leap-of-Thought (CLoT) paradigm, employing associable instruction tuning and explorative self-refinement to significantly enhance LLMs' LoT ability. Further experiments demonstrate the effectiveness and versatility of the proposed CLoT across various creative tasks.  
\end{itemize}
\vspace{0.5cm}

\subsection{The Proposed CLoT is \textcolor{logo2}{not Tailored for} Humor Generation}
\label{sec:tailor}

In this paper, our primary focus is on exploring the Leap-of-Thought (LoT) capability of large language models, a crucial cognitive skill akin to Chain-of-Thought~\cite{wei2022chain,zhang2022automatic,kojima2022large,yao2023tree,long2023large}, rather than humor generation per se. The selection of the Oogiri game as the humor generation task in this study is justified on following three main grounds: 

(1) \underline{\textbf{The Oogiri game serves as an ideal platform for investigating the LLM's LoT ability}}. As discussed in the related works section, the Oogiri game aligns well with the characteristics of LoT, demanding players to think creatively outside the box in response to multimodal information. Moreover, the three primary types within the Oogiri game (including I2T, T2T, and IT2T) also align with the input and output types of multimodal LLMs. Hence, the Oogiri game proves highly suitable for exploring the LoT capability of LLMs; 

(2) \underline{\textbf{The Oogiri game boasts a substantial corpus of manually annotated creative data.}} Due to its widespread popularity on the Internet, the game attracts a large user base generating creative human responses which can constitute an extensive dataset for LoT exploration; 

(3) \underline{\textbf{The Oogiri game facilitates visualization for measuring LoT ability}}. Unlike most association-related cognitive tests ~\cite{beketayev2016scoring,olson2021naming}, the Oogiri game, being inherently a text and image multimodal task, lends itself readily to visualizing LoT ability in a clear and interesting format. Furthermore, this method of showcasing LoT ability, coupled with the amusement factor of the Oogiri game, encourages a broader audience to participate in LoT measurement, ensuring the accuracy of LoT analysis.

\vspace{0.5cm}

\clearpage

\section{More Experimental Results}
\label{sec:moreexpdetail}
In this section, we present additional experimental results to demonstrate the effectiveness of CLoT in enhancing LLM's LoT ability. This includes CLoT's performance across distinct languages (Appendix \ref{sec:different}), diverse responses to single images (Appendix \ref{sec:diver}), and increased Oogiri-style humor generation (Appendix \ref{sec:moreexample}). While humor is \textbf{subjective}, the additional results demonstrate CLoT's LoT capacity of using excellent creative thinking to produce high-quality humor responses.

\subsection{The Performance in Various Languages}
\label{sec:different}

\begin{figure}[h]
  \centering
 \includegraphics[width=0.99\linewidth]{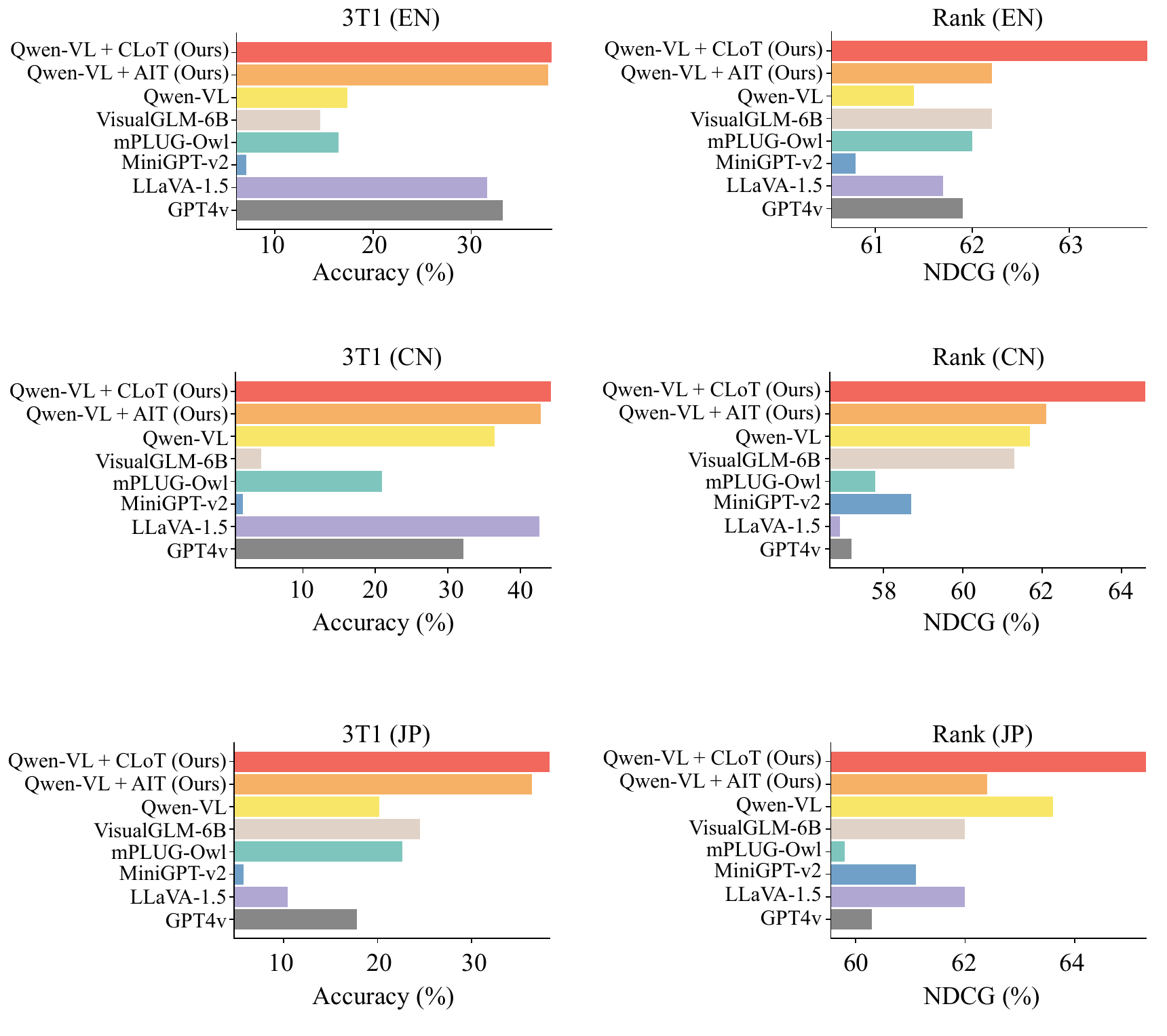}
  \caption{The performance of different LLM for Oogiri game in various languages.}
\label{fig:difflang}
\end{figure}

The results shown in Table 2 in the main text consider all languages together. In this section, we present the performance of all multimodal LLMs individually across different languages, including english (EN), chinese (CN) and japanese (JP). The results are illustrated in Fig.~\ref{fig:difflang}, with Qwen-VL serving as the primary baseline model. It is evident that our proposed CLoT not only significantly enhances the performance of the baseline model but also surpasses other advanced LLMs across various metrics in different languages. This experiment once again underscores the effectiveness of the proposed CLoT.

\subsection{The Diversity Responses of Oogiri Game by LLMs}
\label{sec:diver}

\begin{figure}[h]
  \centering
 \includegraphics[width=0.65\linewidth]{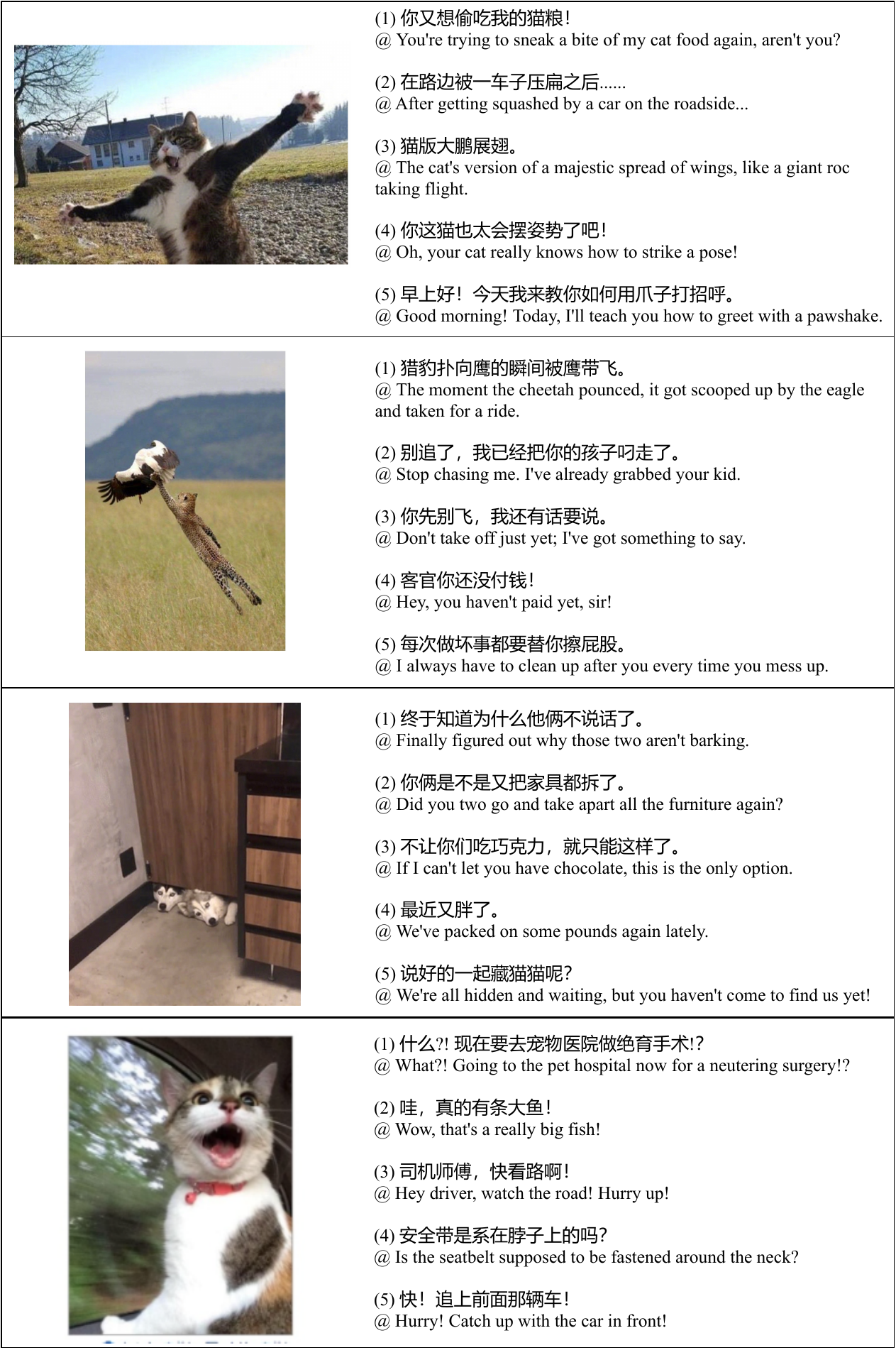}
  \caption{The diversity responses of proposed Creative Leap-of-Thought. ``@" denotes English translations. }
\label{fig:e3}
\end{figure}

To illustrate the creativity of CLoT, we present the diverse responses of Qwen-VL$_{+\text{CLoT}}$ on the same Oogiri game samples, as depicted in Fig.~\ref{fig:e3}. In order to impartially showcase CLoT's creativity, we opt for Chinese Oogiri which has a moderate dataset size in Oogiri-GO, as the limited dataset of English Oogiri may fail to fully capture the advantages of CLoT, and the largest dataset of Japanese Oogiri may overly emphasize CLoT's strengths. To facilitate comprehension for readers of different languages, Fig.~\ref{fig:e3} simultaneously displays the English translations of the Chinese Oogiri responses. However, due to cultural factors and other constraints, the translated content may not entirely convey the intended meaning of the Chinese Oogiri responses. Nevertheless, the diversity of responses in Fig.~\ref{fig:e3} underscores CLoT's ability to engage in divergent thinking and approach challenges from multiple perspectives, showcasing its capacity to think outside the box.

\clearpage
\subsection{More Examples for Oogiri-style Humor Generation}
\label{sec:moreexample}

\begin{figure}[h]
  \centering
 \includegraphics[width=0.75\linewidth]{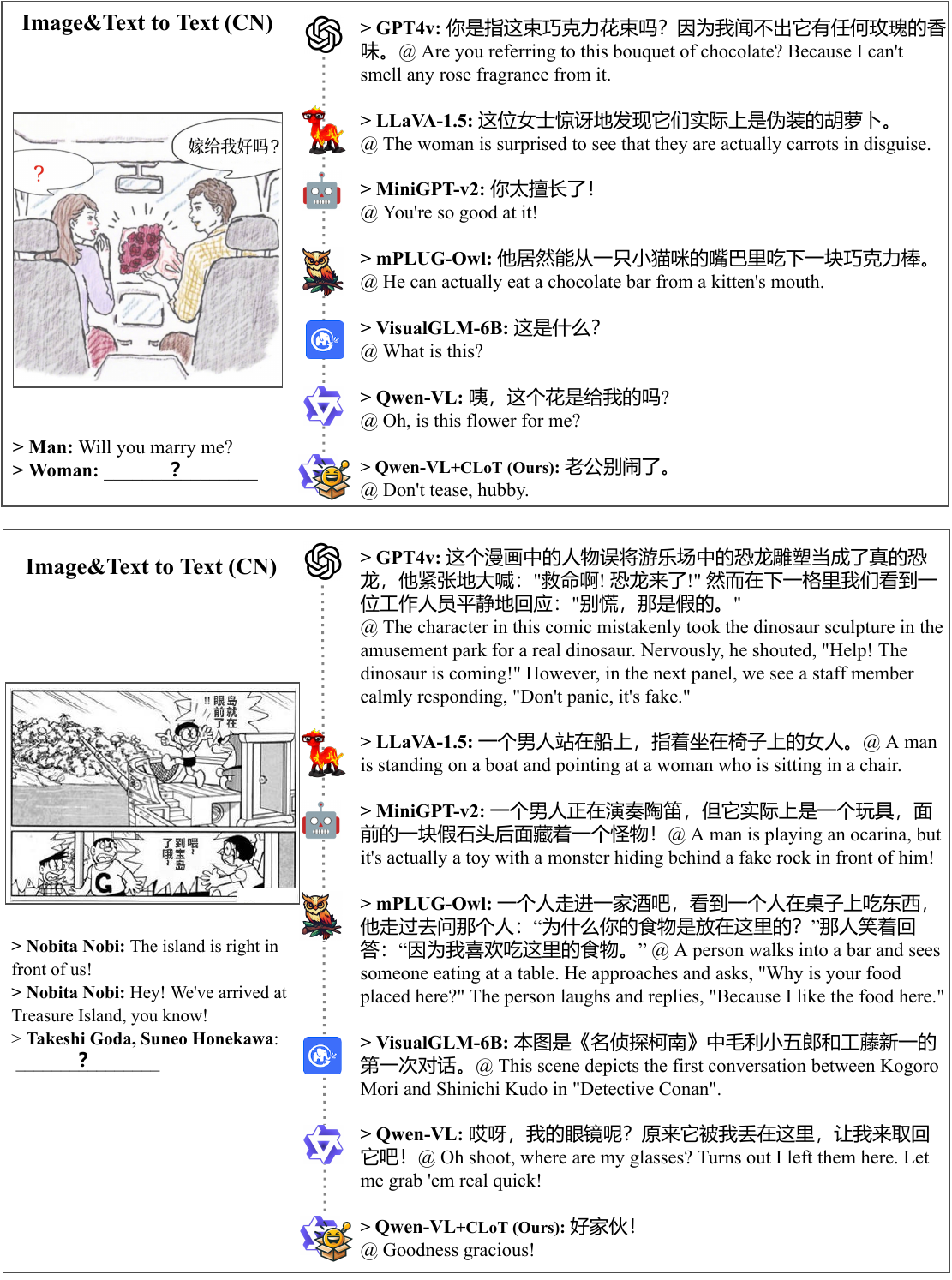}
  \caption{The responses of LLMs in Chinese Oogiri IT2T samples. ``@" denotes English translations. }
\label{fig:e2}
\end{figure}

\begin{figure}[h]
  \centering
 \includegraphics[width=0.80\linewidth]{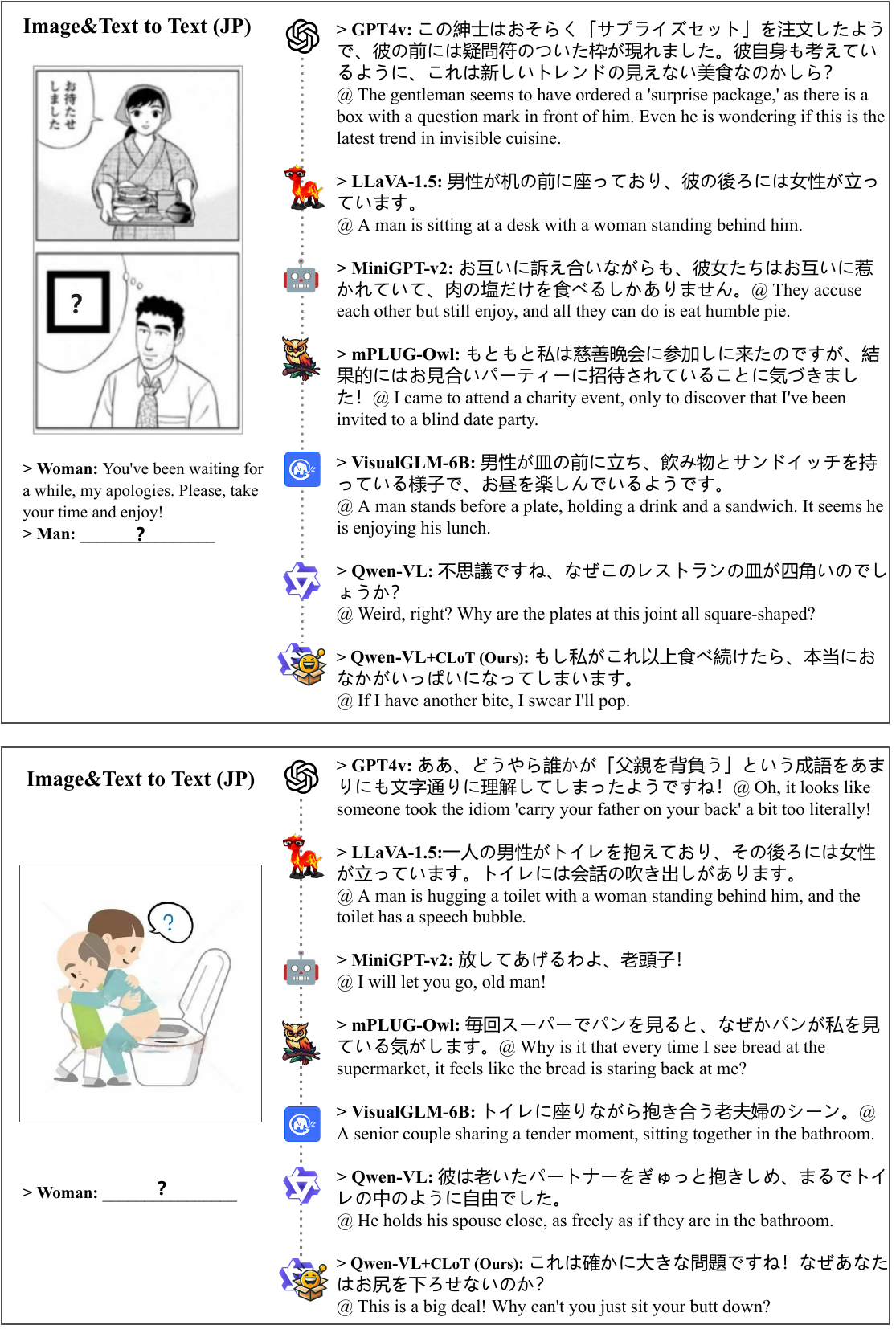}
  \caption{The responses of LLMs in Japanese Oogiri IT2T samples. ``@" denotes translations. }
\label{fig:e1}
\end{figure}

We provide additional examples of humor generation for the multimodal multilingual LLMs mentioned in Table 2 (main text) to illustrate the effectiveness of CLoT. Fig. \ref{fig:e2}, \ref{fig:e1} showcase responses on the task of Image\&Text to Text in Chinese and Japanese, respectively. As English Oogiri data lacks Image\&Text to Text samples, we do not display responses in English for this task (see Appendix \ref{sec:it2teng} for details). Fig. \ref{fig:e7}, \ref{fig:e8}, and \ref{fig:e9} exhibit responses on Image to Text, while Fig. \ref{fig:e6}, \ref{fig:e4}, and \ref{fig:e5} display responses on Text to Text. Finally, we supplement more Qwen-VL$_{+\text{CLoT}}$ responses on Chinese Oogiri samples in Fig. \ref{fig:e10}.

\begin{figure}[h]
  \centering
 \includegraphics[width=0.80\linewidth]{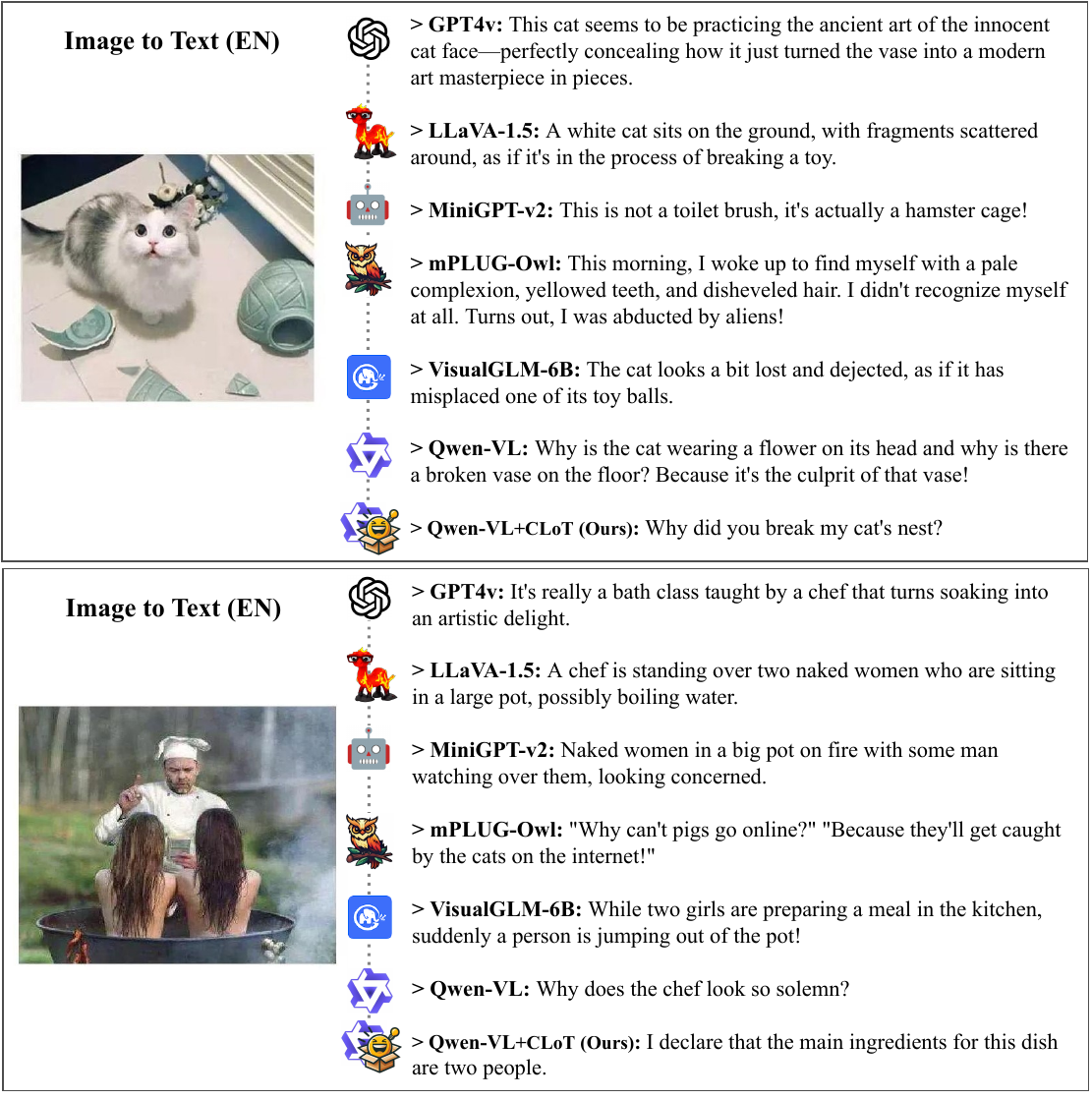}
  \caption{The responses of LLMs in English Oogiri I2T samples.}
\label{fig:e7}
\end{figure}

\begin{figure}[h]
  \centering
 \includegraphics[width=0.80\linewidth]{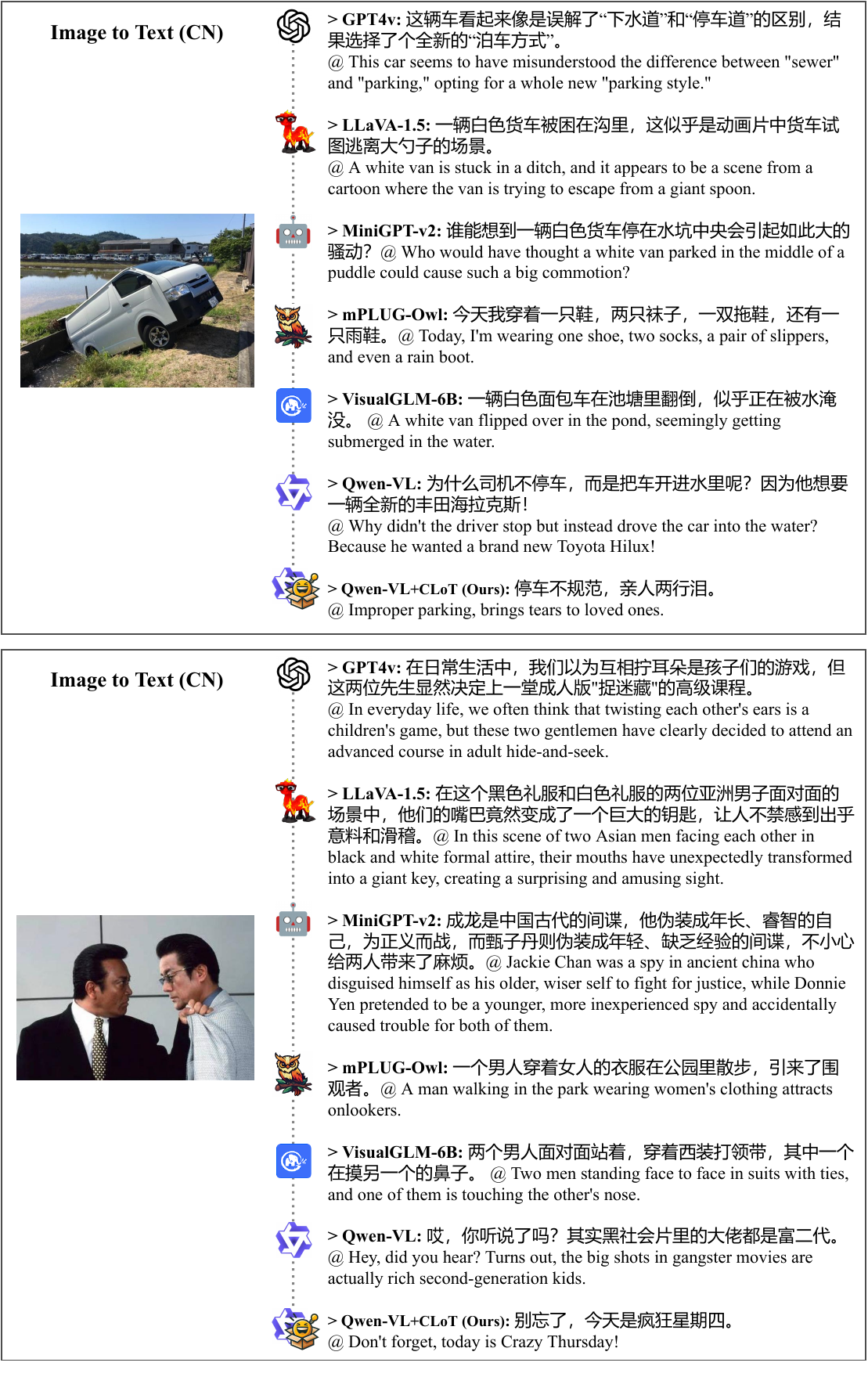}
  \caption{The responses of LLMs in Chinese Oogiri I2T samples. ``@" denotes English translations. }
\label{fig:e8}
\end{figure}

\begin{figure}[h]
  \centering
 \includegraphics[width=0.80\linewidth]{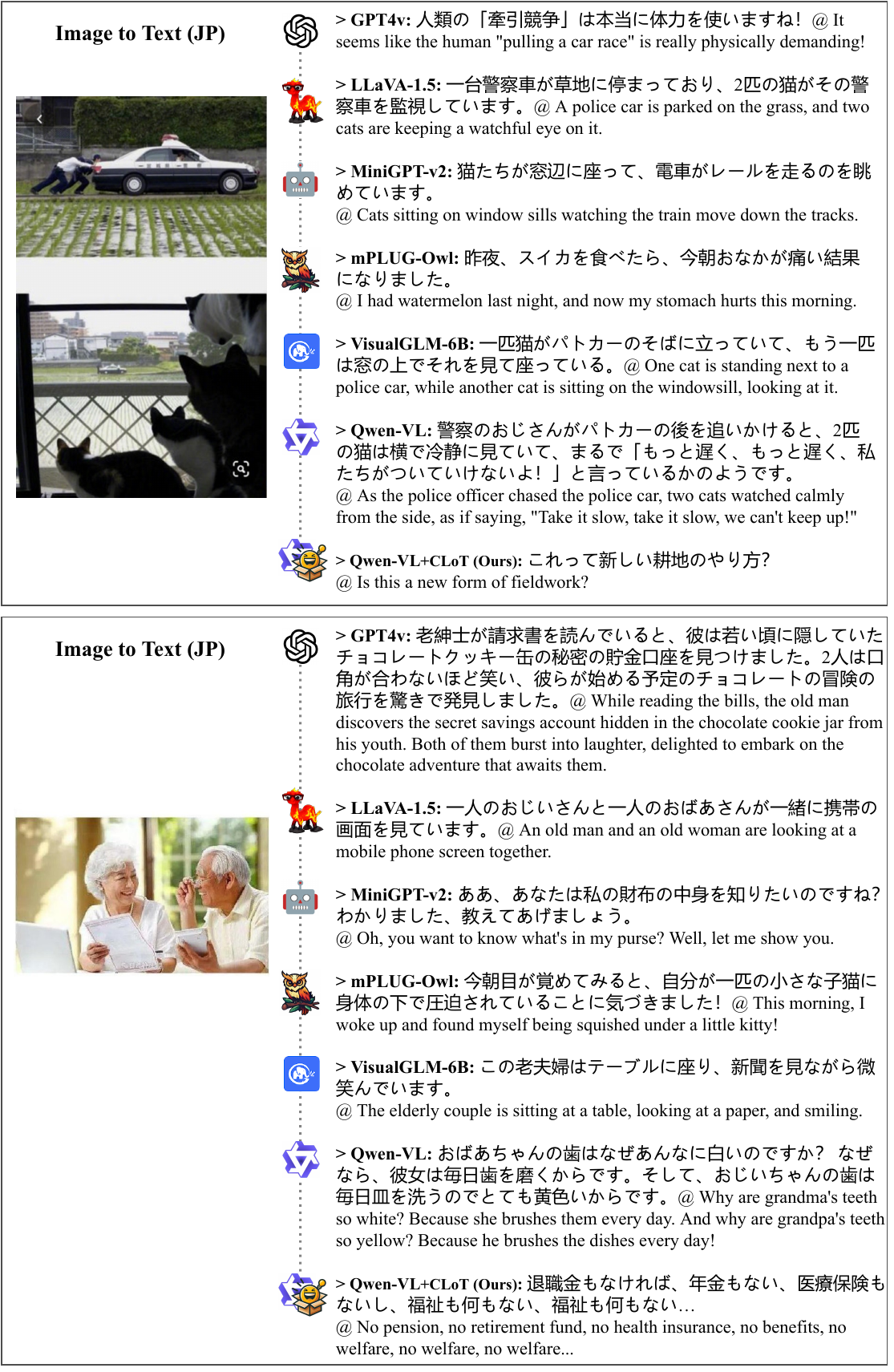}
  \caption{The responses of LLMs in Japanese Oogiri I2T samples. ``@" denotes English translations. }
\label{fig:e9}
\end{figure}

\begin{figure}[h]
  \centering
 \includegraphics[width=0.80\linewidth]{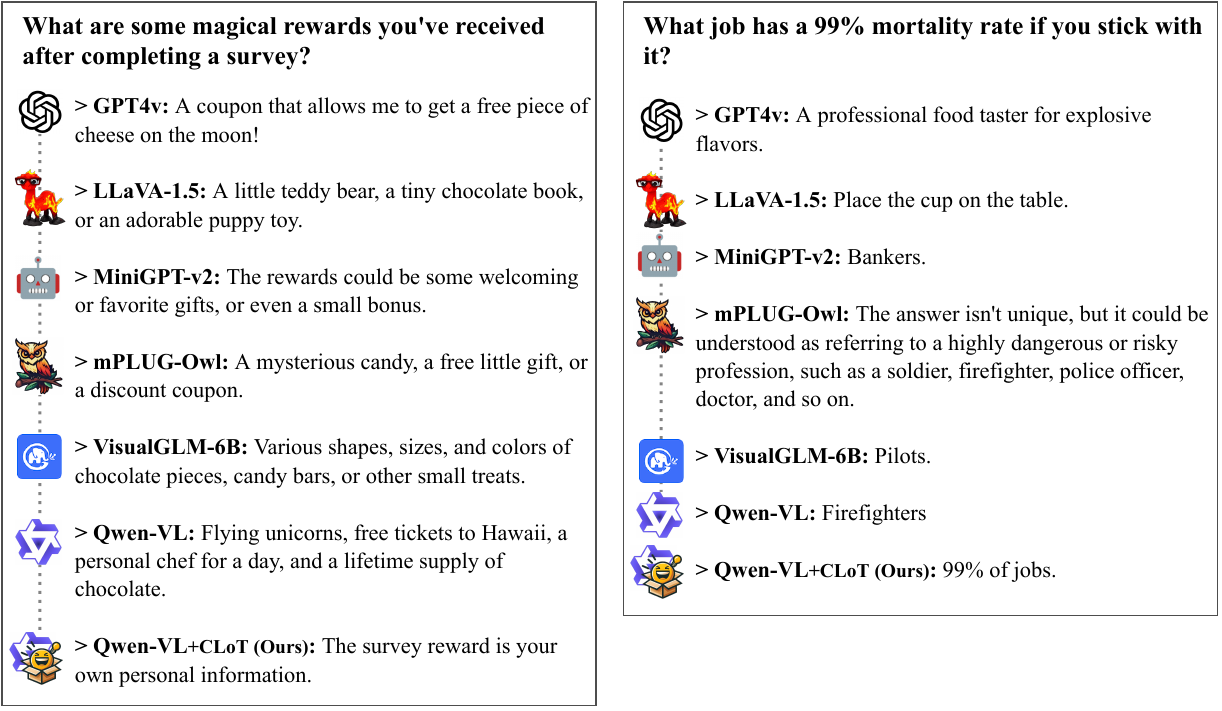}
  \caption{The responses of LLMs in English Oogiri T2T samples.}
\label{fig:e6}
\end{figure}

\begin{figure}[h]
  \centering
 \includegraphics[width=0.80\linewidth]{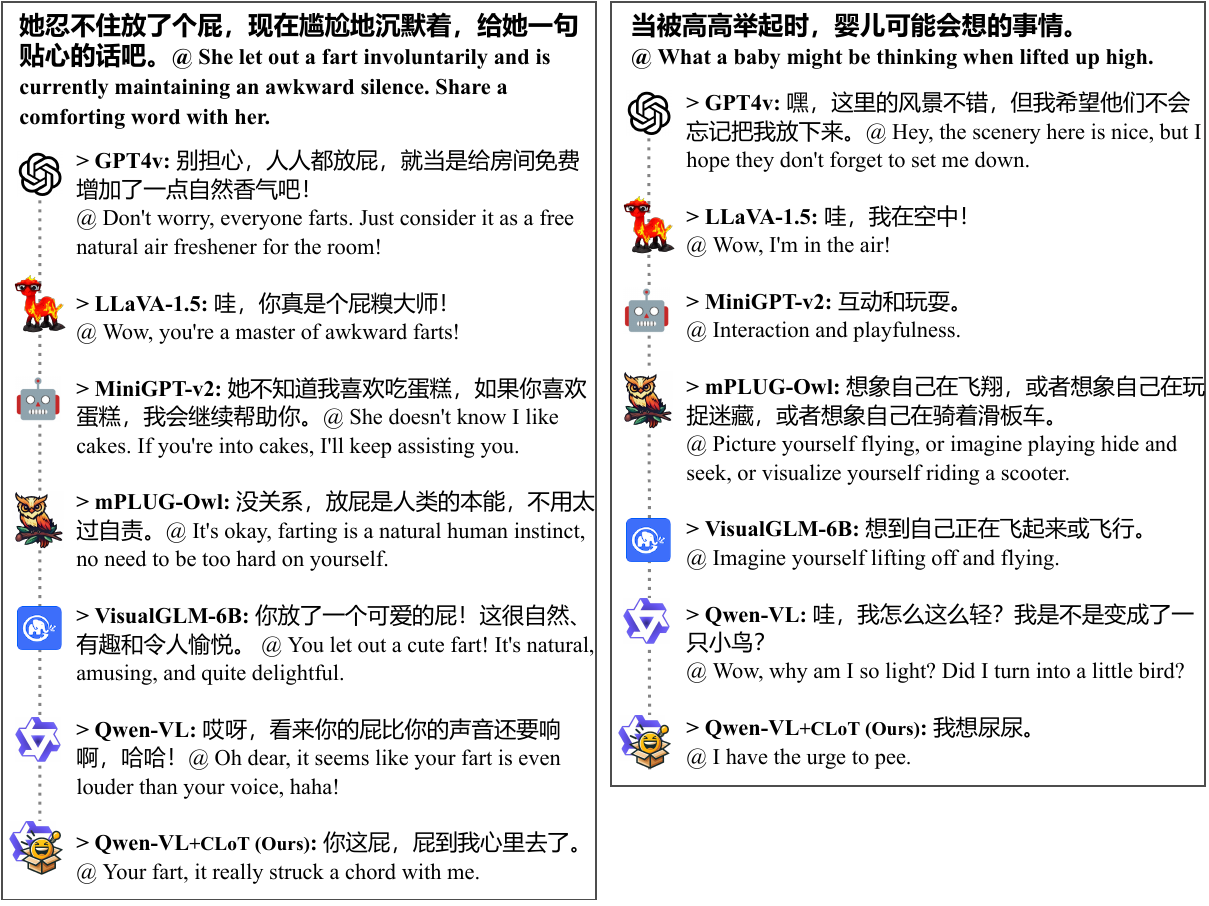}
  \caption{The responses of LLMs in Chinese Oogiri T2T samples. ``@" denotes translations. }
\label{fig:e4}
\end{figure}

\begin{figure}[h]
  \centering
 \includegraphics[width=0.80\linewidth]{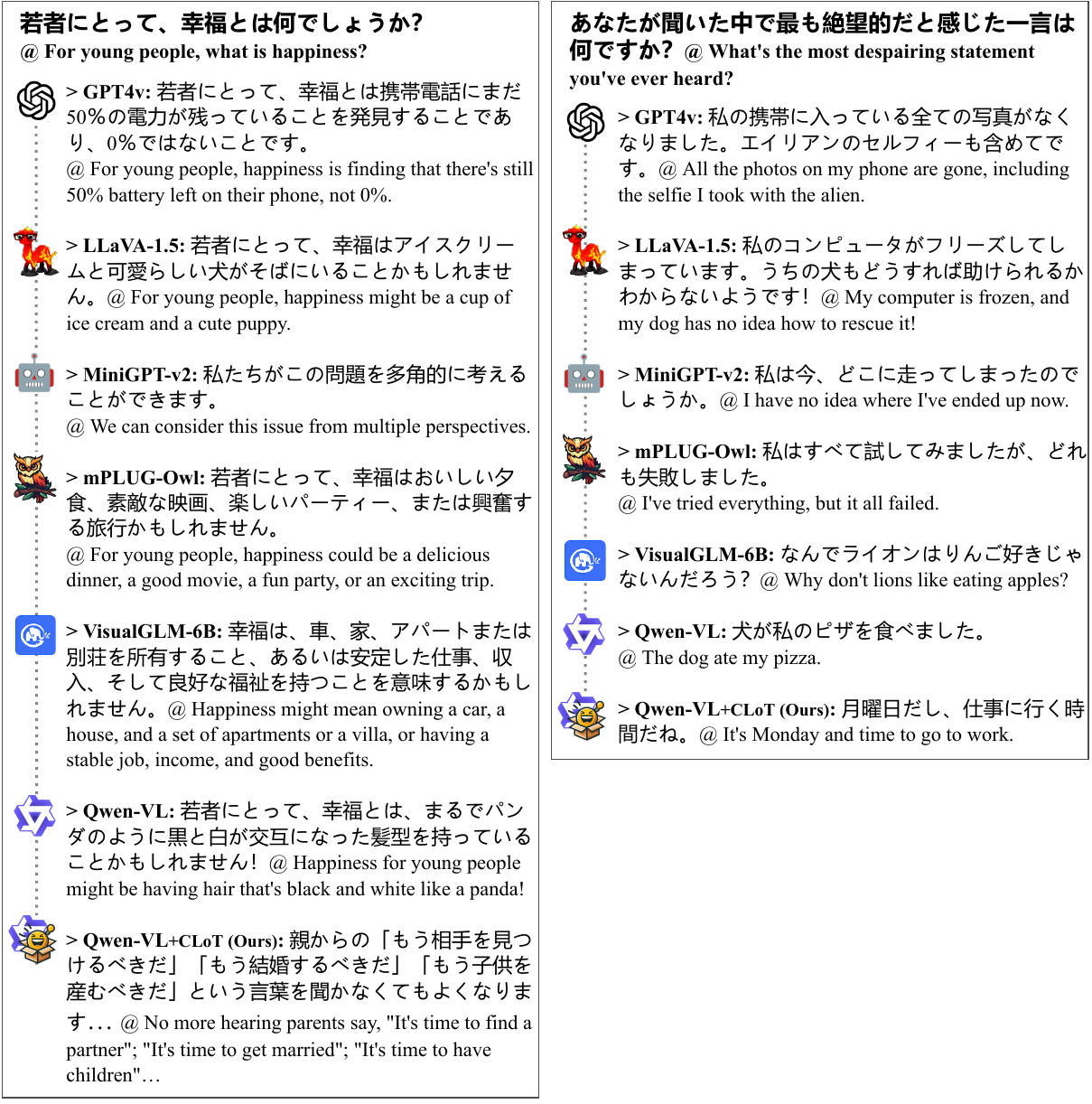}
  \caption{The responses of LLMs in Japanese Oogiri T2T samples. ``@" denotes English translations. }
\label{fig:e5}
\end{figure}

\begin{figure}[h]
  \centering
 \includegraphics[width=0.80\linewidth]{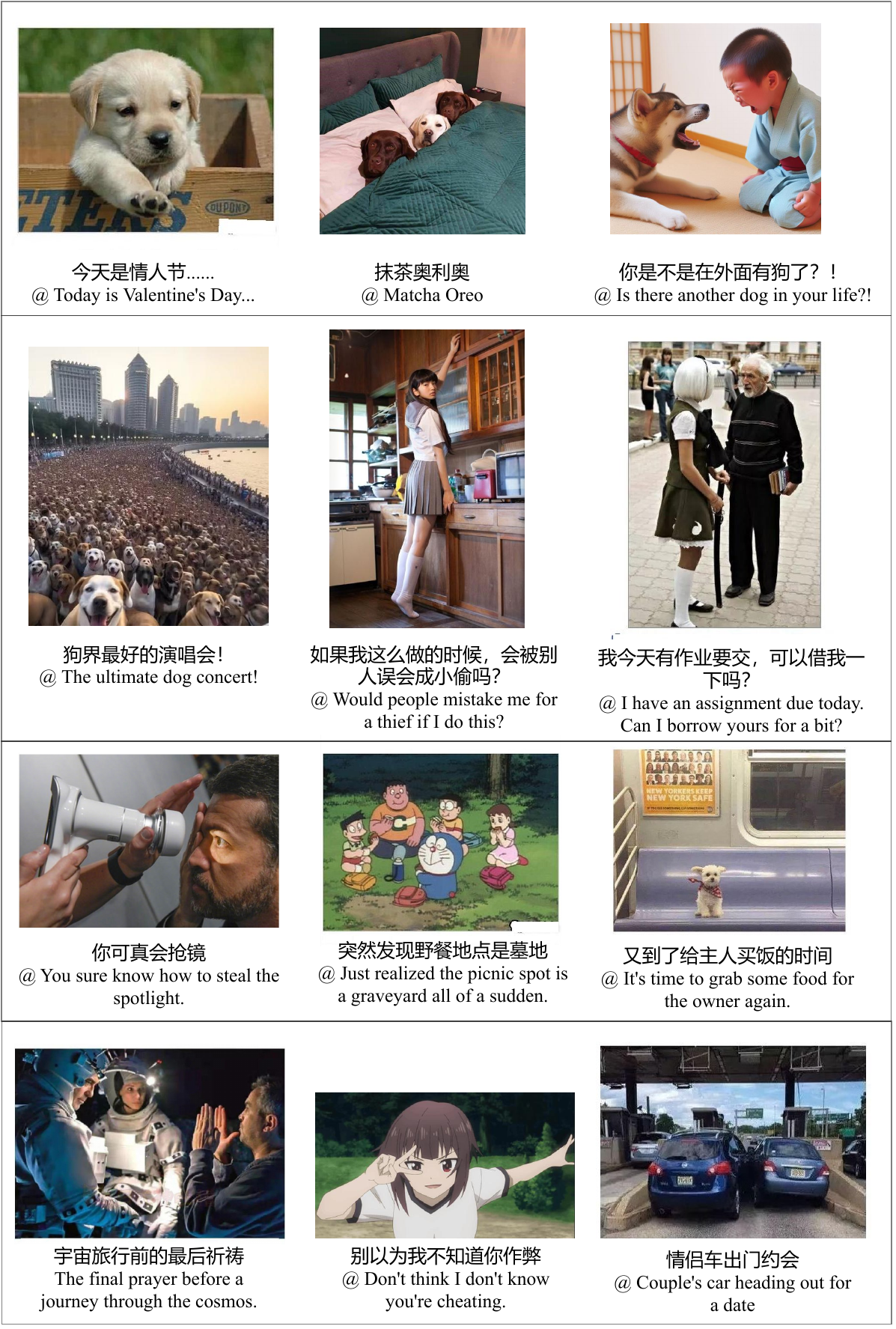}
  \caption{The responses of Qwen-VL$_{+\text{CLoT}}$ in Chinese Oogiri I2T samples. ``@" denotes English translations. }
\label{fig:e10}
\end{figure}

\clearpage
\section{The Construction of Oogiri-GO dataset }
\label{sec:aboutdata}

In this section, we delve into the introduction of data collection and screening for Oogiri-GO. In Appendix \ref{sec:data-collection}, we elucidate the origins of our dataset. Using Bokete as a case study, we expound on the rationale and essential code employed in data crawling. 
Additionally, Appendices \ref{sec:machine-screen} and \ref{sec:manual-screen} provide a detailed breakdown of the procedural steps for machine screening and manual screening, respectively. 

\subsection{Online Data Collection}
\label{sec:data-collection}
We source Oogiri game data from the official  Oogiri game platform, Bokete (https://bokete.jp), and other popular platforms, such as  Twitter (https://twitter.com) and Weibo (https://m.weibo.cn) which also host some Oogiri-game-alike data. Through extensive data collection from different platforms, we gather over 200,000 unfiltered raw samples. Notably, the Bokete website stands out as the preeminent Oogiri game-dedicated platform on the Internet, characterized by the highest data volume and user engagement. Consequently, we select it as a representative case study, providing a comprehensive account of our data acquisition methodology.

\begin{figure}[ht]
  \centering
 \includegraphics[width=0.9\linewidth]{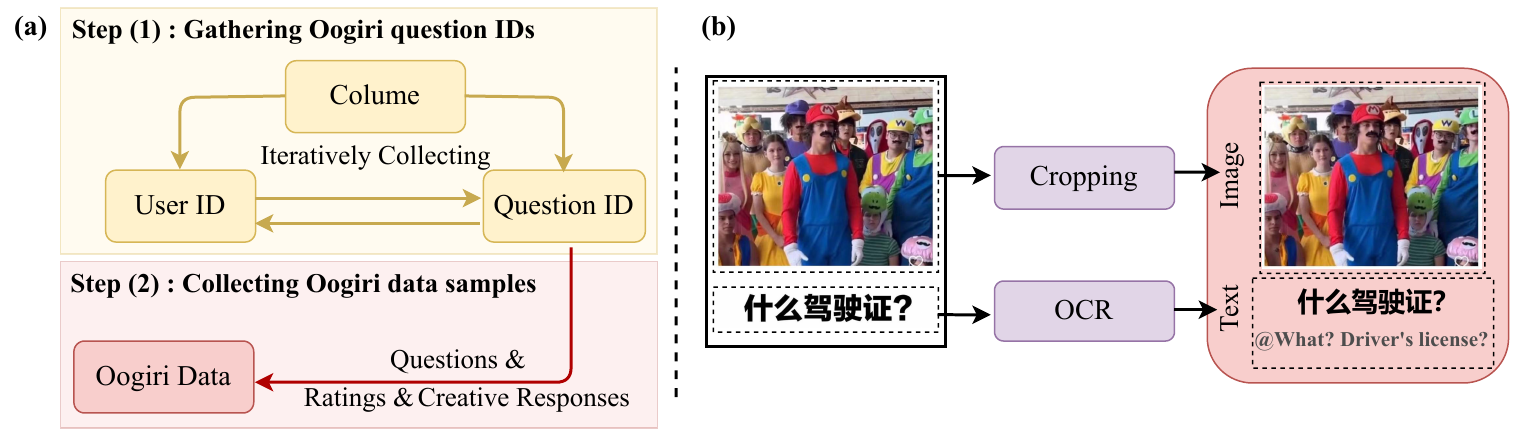}
  \caption{The processing of online data collection. (a) The crawling flow of Bokete website. (b) The image processing of Oogiri data.}
\label{fig:bokete}
\end{figure}

Specifically, as illustrated in Fig. \ref{fig:bokete} (a), the primary approach to crawl the Bokete website involves two key steps: 

(1) \underline{\textbf{Gathering Oogiri question IDs}}. On the Bokete website, an Oogiri question refers to content uploaded by either the official site or users. It exists in the form of images, and even for T2T types, the textual content is embedded in the pictures. Players are tasked with generating creative responses based on these images. Initially, question IDs on the homepage and all corresponding user IDs under each question are preliminarily obtained from various columns, including ``Best", ``Rising", ``Popular" and others. Notably, on each user's homepage, their history of creative responses and ratings is available, allowing us to expand our pool of question IDs from these records. Through this iterative process, we progressively enlarge the pool of both question IDs and user IDs. 

(2) \underline{\textbf{Collecting Oogiri data samples}}. Subsequently, utilizing the gathered question IDs, all creative responses (answers) under a specific question are crawled to compile the Oogiri data. Simultaneously, we record their rating information for the subsequent training of the LLM's discrimination ability in the CLoT framework.

To show the process of online data collection more clearly, we print the core code for both steps below.

\begin{lstlisting}[language=python]
def processing_url(url, page):
    ''' The core code for step (1) Gathering Oogiri question IDs
    Args:
        url (str): basic URL of Bokete, e.g. https://bokete.jp/boke/legend 
        page (int): page number of basic URL, e.g. 1
    '''
    url = f'{url}?page={page}'
    print('processing', url)
    
    # get content of url
    r = requests.get(url)
    r.raise_for_status()
    
    # parse the content and find all hyperlinks <a></a>
    soup = BeautifulSoup(r.text, 'html.parser')
    links = soup.find_all('a', href=True)
    for link in links:
        # find user id
        if "/user/" in link['href']:
            with open('user.txt', 'a') as f:
                f.write(link['href'].split('/')[-1] + '\n')
        # find Oogiri question id
        if "/odai/" in link['href']:
            with open('question.txt', 'a') as f:
                f.write(link['href'].split('/')[-1] + '\n')
\end{lstlisting}

\begin{lstlisting}[language=python]
def processing_odai(odai, page):
    ''' The core code for step (2) Collecting Oogiri data samples
    Args:
        odai (str): question ID, e.g. 6902364
        page (int): page number of question URL, e.g. 1
    '''
    url = f'https://bokete.jp/odai/{odai}?page={page}' 
    print('processing', url)
    
    # get content of url
    r = requests.get(url)
    r.raise_for_status()
    
    # parse the content
    soup = BeautifulSoup(r.text, 'html.parser')
    
    # find the image url of the question
    img = soup.find('a', href=f"/odai/{odai}").find('img')
    link = 'https:' + img.get('src')
    
    # find user id
    links = soup.find_all('a', href=True)
    for link in links:
        if "/user/" in link['href']:
            with open('user.txt', 'a') as f:
                f.write(link['href'].split('/')[-1] + '\n')
    
    # find all answers
    texts = soup.find_all('a', class_='boke-text')
    stars = soup.find_all('div', class_='boke-stars')
    times = soup.find_all('div', class_='boke-information-label')
    for text, star, t in zip(texts, stars, times):
        with open('data.jsonl', 'a') as f:
            f.write(json.dumps({
                'id': text['href'].split('/')[-1],  # id
                'text': text.text,                  # content
                'attitudes_count': star.text,       # rate
                'created_at': t.text,               # creation time
                'pics': {                           # image information
                    'pid': odai,                    # question id
                    'url': link,                    # image link
                }
            }, ensure_ascii=False) + '\n')
\end{lstlisting}

It's worth noting that, unlike the Bokete website, where questions and responses are distinct, on other platforms, data may have questions and responses combined in a single image, as illustrated in Fig. \ref{fig:bokete} (b) \footnote{https://m.weibo.cn/detail/4909366778531862}. In such cases, it is necessary to separate them to construct Oogiri data with a consistent format. Specifically, we utilize PaddleOCR \footnote{https://github.com/PaddlePaddle/PaddleOCR} to recognize text within the main image. Subsequently, leveraging the positional information of the text, we employ image cropping to distinguish the image, resulting in an Oogiri sample with separated questions and responses.

\subsection{Machine Screening by LLM}
\label{sec:machine-screen}

After collecting raw Oogiri data as outlined in Appendix \ref{sec:data-collection}, it is important to acknowledge that the Oogiri game, being a comedy game, may involve responses with biases or other offensive humor. Additionally, since the Oogiri game allows participation from any Internet user, the potential for encountering such issues grows with the game's Internet dissemination. Therefore, effective filtering of the raw data becomes essential. Specifically, to prevent the inclusion of bias, violence, explicit content, offensive language, etc., we employ the multimodal language model Qwen-VL \cite{Qwen-VL} as a checker for the initial screening of the raw data. This screening is performed by constructing safety-checking prompts. The design of the screening template for Qwen-VL is outlined as follows:
\vspace{0.1cm}
\begin{mdframed}[backgroundcolor=gray!8]
\begin{minipage}{\linewidth}
Does the image or text contain content related to \textless Label\textgreater ? Or the combination of image and text shows the metaphor related to \textless Label\textgreater? If so, kindly respond with ``Yes"; otherwise, respond with ``No." 

Here is the text: \textless Text\textgreater
\end{minipage}
\end{mdframed}
\vspace{0.1cm}
where the tag \textless Label\textgreater\ represents the keyword (e.g., violence, explicit content, offensive language, etc.)  and \textless Text\textgreater\ denotes the responses from per collected Oogiri sample.
 To further enhance the effectiveness of safety-checking, we additionally employ the \textless Label\textgreater\ utilized by NudeNet \footnote{https://github.com/notAI-tech/NudeNet}, which includes a substantial number of keywords associated with Not Safe For Work (NSFW) content.
Finally, after machine screening, the number of samples is reduced to about 160,000.

\subsection{Manual Screening}
\label{sec:manual-screen}

Although the majority of inappropriate content is successfully detected from the dataset through machine screening, some subtly metaphorical inappropriate content proves challenging to eliminate entirely. Consequently, aided by translation software, we conducted manual screening to further enhance the quality of the dataset. The criteria for manual screening are consistent with those used in machine screening, involving the removal of content related to the specific \textless Label\textgreater.
The detailed process of manual screening is outlined as two parts. 

(1)\underline{\textbf{ Manual inspection.}} Each sample in the dataset is meticulously examined to determine whether it contains content related to the specified keywords. The inspection involves a careful examination of both images and text to ensure accurate identification and labeling of inappropriate content. 

(2)\underline{\textbf{ Iterative screening.}} To ensure accuracy and consistency in manual screening, we conduct two rounds of iterative manual screening. Each round involves different individuals to minimize the impact of subjective judgments and enhance the reliability of the dataset. Following manual screening, we successfully further reduce the presence of inappropriate content in the dataset, refining the sample count to more than 130,000. The introduction of manual screening contributes to ensuring a high-quality dataset and more sensitive detection of inappropriate content.

\clearpage
\section{Experimental Details}
\label{sec:expdetails}

\subsection{The Details of Implementation Details }
\textbf{Metrics. }
In this paper, we delve into experiments that include choice and ranking questions inspired by the humor benchmarks in \cite{hessel2023androids}. Additionally, we conduct a user study to directly evaluate the effectiveness of humor generation, along with other creative tasks such as the Cloud Guessing Game (CGG) and the Divergent Association Task (DAT). Subsequently, individual metrics for each of these experiments will be provided.

(1) For the \textit{choice questions},
we utilize classification accuracy as the evaluation metric. Specifically,  the accuracy of LLMs is calculated by dividing the number of correctly answered questions by the total number of questions. 
 
(2)  For the \textit{ranking questions}. 
we adopt the widely used ranking metric, i.e., Normalized Discounted Cumulative Gain (NDCG) \cite{radlinski2010comparing}. We adpot top-1 accuracy as the positions at the top of rank lists are more significant in ranking scenarios \cite{tang2018ranking}.

(3)  For the \textit{user study}, we conduct a user survey, tallying the total number of votes received by various LLMs across different categories of Oogiri. Subsequently, we calculat the percentage of votes each LLM garnered in relation to its overall vote count for different types. see Appendix \ref{sec:user} for more details.

(4) 
For the \textit{other creative tasks}, we employed classification accuracy and average semantic distance (ASD) as metrics for the CGG and DAT tasks, respectively . ASD represents the average semantic distance of all test examples, where the semantic distance for each test example is calculated based on the ten words following the completion of each choice question. see Appendix \ref{sec:othercreative} for more details.

\noindent\textbf{Hyperparameters of Associable Instruction Tuning.}
For ``Image" condition, it relies on the type of Oogiri game, e.g., being the image embeddings in I2T game and empty in T2T type. For the ``condition" option, it's set to empty with a probability of $\rho_c$,  and otherwise is randomly set as one noun in ``task-specific responses". We set the value of $\rho_c$ to 0.50. This setting is driven by the fact that training LLMs to perform associable generation assists in the remote association of self-refinement, and unconditionally controlling leap-of-thought generation is the capability we aim for the model to acquire.

\noindent\textbf{Hyperparameters of Explorative Self-Refinement.}
During explorative remote association,  we generate $n$ weakly-associated conditions $\{\Cm_i\}_{i=1}^n$. These conditions can either be empty with a probability $\rho=0.5$ to give freedom to LLM, or uniformly randomly sampled from the noun set $\Sm$ to enforce LLM to build connections between different concepts. Next, we add the condition $\Cm_i$ into user-input $\Imm$,  and feed $\Imm$ into the LLM to generate a humor candidate $\Rm_i$. Repeating this process with different conditions $\Cm_i$ can generate  a total of $n$ candidates  $\{\Rm_i\}_{i=1}^n$. 
We set the value of $n$ to 5, aiming not only to control the difficulty of ranking for reliable results but also to align with the number of options in the discrimination during associable instruction tuning. 
Then the LLM ranks these candidates by its discriminative ranking ability learned in Sec. 4.1 (main text). Next, it mixes the top-2 candidates with the ground truth responses, and selects the top-1 as the final response.

Here we not only incorporate ranking but also introduce a selection process to achieve explorative remote association. This decision is based on our experimental results, as demonstrated in the experimental section, indicating that the accuracy of LLMs tends to increase with a decrease in the number of choices for choice questions. Directly having LLMs choose an option from a pool of $n$ candidates poses a significant challenge. Hence, we design a two-step process involving ranking followed by selection. Furthermore, the choice of selecting the Top-2 candidates from the ranking results is intended to ensure the accuracy of LLMs in completing choice questions.

\noindent\textbf{Hyperparameters of Training.}
We use the official code of Qwen-VL~\cite{Qwen-VL} and  CogVLM~\cite{wang2023cogvlm} for implementation and training. All models are trained utilizing 8 Nvidia A100 (40G) GPUs. The training and hyperparameters for each model are specified as follows. 

(1) Qwen-VL$_{\text{+CLoT}}$ is trained using AdamW optimizer with $\beta_{1}=0.9, \beta_{2}=0.95, eps=1e^{-8}$. We set the learning rate to $1e^{-5}$ and use a weight decay of $1e^{-1}$. The training process uses a batch size of 64. LoRA in Qwen-VL has a rank of 64, a normalization parameter of 16, and a dropout rate of 0.05.

(2) CogVLM-17B$_{\text{+CLoT}}$ is trained using AdamW optimizer with $\beta_{1}=0.9, \beta_{2}=0.95, eps=1e^{-8}$. We set the learning rate to $1e^{-5}$ and use a weight decay of $5e^{-2}$. The training process uses a batch size of 128. LoRA in CogVLM-17B has a rank of 10, a normalization parameter of 1, and a dropout rate of 0.00.

\subsection{The Details of Instruction Templates}

\begin{figure}[h]
    \centering
    \includegraphics[width=0.45\linewidth]{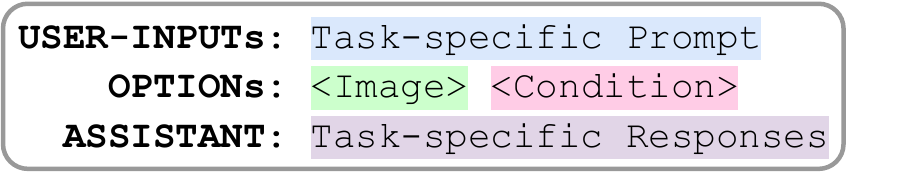}
    \caption{The LoT-oriented instruction templates.}
    \label{fig:template2}
\end{figure}

After completing data collection and screening, the next step is to transform the collected Oogiri data into instruction-tuning data that will be utilized to train models. We design some LoT-oriented instruction templates to transform the Oogiri-GO dataset into instruction tuning data, and then train LLM to achieve associable generation and discrimination abilities. Our templates primarily comprise two components in Fig.~\ref{fig:template2}: task-specific prompt and response.  For different abilities,  the templates need some special design.  
In this section, we will elaborate on the details of instruction templates for each task.

\noindent\textbf{Instruction Templates of Image to Text. }Based on Fig.~\ref{fig:template2}, we can categorize the instruction templates for Image to Text into the following four types:

\vspace{0.1cm}
\begin{mdframed}[backgroundcolor=gray!8]
\begin{minipage}{\linewidth}
\textbf{Original Instruction}

\sethlcolor{prompt}\hl{Based on the image, think of a sentence that is unexpected and humorous. Let's think outside the box. A satisfactory response is}

\sethlcolor{image}\hl{Image: \textless Image\textgreater}

\sethlcolor{response}\hl{\textless Response\textgreater}\\

\textbf{Instruction with Condition}

\sethlcolor{prompt}\hl{Please carefully understand the image and give an answer that contains conditional words and is surprising and funny. Let's think outside the box. A surprising and funny answer containing conditional word is}

\sethlcolor{condition}\hl{Condition: \textless Condition\textgreater}

\sethlcolor{image}\hl{Image: \textless Image\textgreater}

\sethlcolor{response}\hl{\textless Response\textgreater}\\

\textbf{Instruction for Ranking}

\sethlcolor{prompt}\hl{Please evaluate the degree of unexpected and humorous effect when each of the option contents is combined with the image. 

Options: \\
    A. \textless Content A\textgreater\\
    B. \textless Content B\textgreater\\
    C. \textless Content C\textgreater\\
    D. \textless Content D\textgreater\\
    E. \textless Content E\textgreater\\
Response Format: Please respond in the format of ranking the humorousness of the options from high to low, for example, ``1. A. xxx. 2. B. xxx. 3. C. xxx. 4. D. xxx. 5. E. xxx.". Be sure to rank all five options.

Let's think outside the box. The result of ranking the options from most surprising and funny to least is
}

\sethlcolor{image}\hl{Image: \textless Image\textgreater}

\sethlcolor{response}\hl{\textless Response\textgreater}\\

\textbf{Instruction for 3T1 Selection}

\sethlcolor{prompt}\hl{Please select the option that, when combined with the image, creates an unexpected and humorous effect. Only one option meets the requirements. 

Options: \\
    A. \textless Content A\textgreater\\
    B. \textless Content B\textgreater\\
    C. \textless Content C\textgreater\\  
Response Format: Please respond in the format of ``Option id. Option content", for example, ``A. xxx".

Let's think outside the box. The satisfactory option is}

\sethlcolor{image}\hl{Image: \textless Image\textgreater}

\sethlcolor{response}\hl{\textless Response\textgreater}\\

\end{minipage}
\end{mdframed}
\vspace{0.1cm}
where the tags \textless Image\textgreater, \textless Response\textgreater, \textless Condition\textgreater\ and \textless Content $X$\textgreater\ serve as placeholders for inserting the embeddings of visual image, the text response, the text condition, and the text option content. 
The condition of instruction with condition is from the nouns of ground truth response, and the candidate options of instruction for ranking are from the Oogiri data with multiple answers. 
Besides, we illustrate the instructions for selection taking 3T1 selection as an example. For other types of selection instructions, only minor modifications to the number of options and quantifiers are needed.

We set the number of original instruction templates for each task to three. This decision is rooted in the precise requirements of the three tasks in Oogiri-GO, which is that we aim for LLMs to generate unexpected and humorous content based on given images or texts. Consequently, a large number of prompts is unnecessary for robust generalization across these tasks and our experiments support this observation. In the Image to Text task, we compare the effects of using three and fifty prompt templates and the fifty prompt templates are obtained by rewriting the original three templates using Qwen-14B~\cite{bai2023qwen}. The experimental results demonstrate that the number of prompt templates does not significantly impact the performance of LLMs. 

\noindent\textbf{Instruction Templates of Text to Text. }
Instruction templates for Text to Text are very similar to those for Image to Text, and can also be categorized into the following four types:

\vspace{0.1cm}
\begin{mdframed}[backgroundcolor=gray!8]
\begin{minipage}{\linewidth}
\textbf{Original Instruction}

\sethlcolor{prompt}\hl{Please carefully understand the provided question and come up with a surprising and humorous response. }

\sethlcolor{prompt}\hl{Question: \textless Question\textgreater}

\sethlcolor{prompt}\hl{Let's think outside the box. A satisfactory response is}

\sethlcolor{response}\hl{\textless Response\textgreater}\\

\textbf{Instruction with Condition}

\sethlcolor{prompt}\hl{Please carefully understand the question and give an answer that contains conditional words and is surprising and funny.}
\sethlcolor{prompt}\hl{Question: \textless Question\textgreater}

\sethlcolor{prompt}\hl{Let's think outside the box. A surprising and funny answer containing conditional word is}

\sethlcolor{condition}\hl{Condition: \textless Condition\textgreater}

\sethlcolor{response}\hl{\textless Response\textgreater}\\

\textbf{Instruction for Ranking}

\sethlcolor{prompt}\hl{Please evaluate the degree of unexpected and humorous effect when each of the option contents is combined with the question.

Question: \textless Question\textgreater\\
Options: \\
    A. \textless Content A\textgreater\\
    B. \textless Content B\textgreater\\
    C. \textless Content C\textgreater\\
    D. \textless Content D\textgreater\\
    E. \textless Content E\textgreater\\
Response Format: Please respond in the format of ranking the humorousness of the options from high to low, for example, ``1. A. xxx. 2. B. xxx. 3. C. xxx. 4. D. xxx. 5. E. xxx.". Be sure to rank all five options.

Let's think outside the box. The result of ranking the options from most surprising and funny to least is
}

\sethlcolor{response}\hl{\textless Response\textgreater}\\

\textbf{Instruction for 3T1 Selection}

\sethlcolor{prompt}\hl{Please select the option that, when combined with the question, creates an unexpected and humorous effect. Only one option meets the requirements. 

Question: \textless Question\textgreater\\
Options: \\
    A. \textless Content A\textgreater\\
    B. \textless Content B\textgreater\\
    C. \textless Content C\textgreater\\  }
\end{minipage}
\end{mdframed}
\vspace{0.1cm}

\vspace{0.1cm}
\begin{mdframed}[backgroundcolor=gray!8]
\begin{minipage}{\linewidth}

\sethlcolor{prompt}\hl{Response Format: Please respond in the format of ``Option id. Option content", for example, ``A. xxx".

Let's think outside the box. The satisfactory option is}

\sethlcolor{response}\hl{\textless Response\textgreater}\\

\end{minipage}
\end{mdframed}
\vspace{0.1cm}
where the tag \textless Question\textgreater\ denotes the text question of Oogiri data.

\noindent\textbf{Instruction Templates of Image\&Text to Text. }
The instruction templates for Image\&Text to Text are similar to those of the other two tasks, but due to the unique nature of Image\&Text to Text, we incorporate a special character [MASK] into the templates. The instruction templates for Image\&Text to Text are as follows:

\vspace{0.1cm}
\begin{mdframed}[backgroundcolor=gray!8]
\begin{minipage}{\linewidth}
\textbf{Original Instruction}

\sethlcolor{prompt}\hl{In this image, there are sections of text that need to be completed, and the content to fill in is denoted by [MASK]. Let's think outside the box and complete the [MASK] to make the response unexpectedly funny. A satisfactory response is}

\sethlcolor{image}\hl{Image: \textless Image\textgreater}

\sethlcolor{response}\hl{\textless Response\textgreater}\\

\textbf{Instruction with Condition}

\sethlcolor{prompt}\hl{In this image, there are sections of text that need to be completed, and the content to fill in is denoted by [MASK]. Let's think outside the box and complete the [MASK] with a response that contains conditional words and is surprising and funny. A surprising and funny response containing conditional word is}

\sethlcolor{condition}\hl{Condition: \textless Condition\textgreater}

\sethlcolor{image}\hl{Image: \textless Image\textgreater}

\sethlcolor{response}\hl{\textless Response\textgreater}\\

\textbf{Instruction for Ranking}

\sethlcolor{prompt}\hl{In this image, there are sections of text that need to be completed, and the content to fill in is denoted by [MASK]. Please evaluate the degree of unexpected and humorous effect when the options are the content of the [MASK]. 

Options: \\
    A. \textless Content A\textgreater\\
    B. \textless Content B\textgreater\\
    C. \textless Content C\textgreater\\
    D. \textless Content D\textgreater\\
    E. \textless Content E\textgreater\\
Response Format: Please respond in the format of ranking the humorousness of the options from high to low, for example, ``1. A. xxx. 2. B. xxx. 3. C. xxx. 4. D. xxx. 5. E. xxx.". Be sure to rank all five options.

Let's think outside the box. The result of ranking the options from most surprising and funny to least is
}

\sethlcolor{image}\hl{Image: \textless Image\textgreater}

\sethlcolor{response}\hl{\textless Response\textgreater}\\

\textbf{Instruction for 3T1 Selection}

\sethlcolor{prompt}\hl{In this image, there are sections of text that need to be completed, and the content to fill in is denoted by [MASK]. Please select the option that, creates an unexpected and humorous effect when being the content of the [MASK]. Only one option meets the requirements. 

Options: \\
    A. \textless Content A\textgreater\\
    B. \textless Content B\textgreater\\
    C. \textless Content C\textgreater\\  
Response Format: Please respond in the format of ``Option id. Option content", for example, ``A. xxx".

Let's think outside the box. The satisfactory option is}

\sethlcolor{image}\hl{Image: \textless Image\textgreater}

\sethlcolor{response}\hl{\textless Response\textgreater}\\

\end{minipage}
\end{mdframed}
\vspace{0.1cm}

From Table 1 in the main text, it is evident that the data volume for the IT2T category is significantly lower compared to the other two Oogiri game types. Consequently, there is a need to augment more data to enhance instruction tuning for IT2T.
It is noteworthy that IT2T can be regarded as a form of masked language modeling (MLM) task~\cite{sinha2021masked,salazar2019masked,bitton2021data}. Therefore, we contemplate achieving this objective by constructing MLM tasks for the I2T and T2T data types. The details are as follows:
\vspace{0.1cm}
\begin{mdframed}[backgroundcolor=gray!8]
\begin{minipage}{\linewidth}
\textbf{Mask Instruction for I2T}

\sethlcolor{prompt}\hl{Please carefully understand the provided image and complete the answer by replacing the [MASK] part to make the answer unexpectedly funny. 

Answer: \textless Answer with [MASK]\textgreater \\
Let's think outside the box. The content of [MASK] is
}

\sethlcolor{image}\hl{Image: \textless Image\textgreater}

\sethlcolor{response}\hl{\textless Response\textgreater}\\

\textbf{Mask Instruction for T2T}

\sethlcolor{prompt}\hl{Please carefully understand the provided question and complete the answer by replacing the [MASK] part to make the answer unexpectedly funny. 

Question: \textless Question\textgreater \\
Answer: \textless Answer with [MASK]\textgreater \\
Let's think outside the box. The content of [MASK] is
}

\sethlcolor{response}\hl{\textless Response\textgreater}\\

\end{minipage}
\end{mdframed}
\vspace{0.1cm}
Here, we probabilistically replace nouns or verb phrases from the Oogiri answers with [MASK], and use the replaced Oogiri answer as \textless Answer with [MASK]\textgreater.

\subsection{The Details of Tuning by LoRA}

LoRA~\cite{hu2021lora} is a widely employed method for fine-tuning LLMs. It effectively reduces the number of trainable parameters by learning pairs of rank-decomposition matrices while maintaining the original weights in a frozen state. 
LoRA currently stands out as a superior adaptation method. Hence, we train LoRA for the LLMs with the associable instruction data. 

The code snippets below illustrate the insertion points for LoRA during the training of Qwen-VL~\cite{Qwen-VL}. The first code snippet demonstrates how to insert LoRA into the textual module of Qwen-VL, while the second code snippet shows how to insert LoRA into the visual module of Qwen-VL.
\begin{lstlisting}[language=python]
# add LoRA to the textual module of Qwen-VL
QWenLMHeadModel(
  (transformer): QWenModel(
    (wte): Embedding(151936, 4096)
    (drop): Dropout(p=0.0, inplace=False)
    (rotary_emb): RotaryEmbedding()
    (h): ModuleList(
      (0-31): 32 x QWenBlock(
        (ln_1): RMSNorm()
        (attn): QWenAttention(
          (c_attn): Linear(in_features=4096, out_features=12288, bias=True) # + LoRA
          (c_proj): Linear(in_features=4096, out_features=4096, bias=False) # + LoRA
          (attn_dropout): Dropout(p=0.0, inplace=False)
        )
        (ln_2): RMSNorm()
        (mlp): QWenMLP(
          (w1): Linear(in_features=4096, out_features=11008, bias=False)
          (w2): Linear(in_features=4096, out_features=11008, bias=False)
          (c_proj): Linear(in_features=11008, out_features=4096, bias=False)
        )
      )
    )
......
\end{lstlisting}

\begin{lstlisting}[language=python]
# add LoRA to the visual module of Qwen-VL
QWenLMHeadModel(
  (transformer): QWenModel(
    ......
    (visual): VisionTransformer(
      (conv1): Conv2d(3, 1664, kernel_size=(14, 14), stride=(14, 14), bias=False)
      (ln_pre): LayerNorm((1664,), eps=1e-06, elementwise_affine=True)
      (transformer): TransformerBlock(
        (resblocks): ModuleList(
          (0-47): 48 x VisualAttentionBlock(
            (ln_1): LayerNorm((1664,), eps=1e-06, elementwise_affine=True)
            (ln_2): LayerNorm((1664,), eps=1e-06, elementwise_affine=True)
            (attn): VisualAttention(
              (in_proj): Linear(in_features=1664, out_features=4992, \
              bias=True)  # + LoRA
              (out_proj): Linear(in_features=1664, out_features=1664, \
              bias=True)  # + LoRA
            )
            (mlp): Sequential(
              (c_fc): Linear(in_features=1664, out_features=8192, bias=True) # + LoRA
              (gelu): GELU(approximate='none')
              (c_proj): Linear(in_features=8192, out_features=1664, bias=True)
            )
          )
        )
      )
      (attn_pool): Resampler(
        (kv_proj): Linear(in_features=1664, out_features=4096, bias=False)
        (attn): MultiheadAttention(
          (out_proj): NonDynamicallyQuantizableLinear(in_features=4096, \
          out_features=4096, bias=True)  # + LoRA
        )
        (ln_q): LayerNorm((4096,), eps=1e-06, elementwise_affine=True)
        (ln_kv): LayerNorm((4096,), eps=1e-06, elementwise_affine=True)
      )
      (ln_post): LayerNorm((4096,), eps=1e-06, elementwise_affine=True)
    )
  )
  (lm_head): Linear(in_features=4096, out_features=151936, bias=False)
)
\end{lstlisting}

To investigate the appropriate insertion strategy for LoRA, we conduct three sets of associable instruction tuning experiments using Oogiri-GO I2T data. LoRA is inserted separately into the textual, visual, and both textual and visual modules of Qwen-VL. Experimental results indicate that, based on the 3T1 metric, the accuracy of LoRA insertion solely into the textual module (38.8) surpasses the performance of simultaneous insertion into both textual and visual modules (37.0), while the accuracy is lowest when LoRA is inserted only into the visual module (25.4). Therefore, we exclusively train LoRA in the textual module of Qwen-VL.

\subsection{The Extraction of Weakly-associated Conditions}
The core of explorative remote association is to prompt the LLM to generate a diverse array of creative responses under weakly-associated conditions. To implement this, we extract a set of object nouns from the text in the Oogiri-GO training data. In this section, we provide the details of the extraction of weakly-associated conditions. 

We initiate by extracting a set of nouns from the responses in data of Oogiri-GO. For this, we employ NLTK \cite{bird2009natural}, Jieba \footnote{https://github.com/fxsjy/jieba}, and Janome \footnote{https://github.com/mocobeta/janome} for various languages, considering the performance differences of different part-of-speech analysis tools across languages. 

Specifically, we utilize NLTK, a suite of open source Python modules, data sets, and tutorials supporting research and development in natural language processing, to extract nouns from English text and the core code snippet is presented below:
\begin{lstlisting}[language=python]
from nltk import word_tokenize, pos_tag

def extract_nouns(text):
    tokens = word_tokenize(text)
    tagged_words = pos_tag(tokens)
    conditions = [word for word, pos in tagged_words if pos.startswith('N')]
    return conditions
\end{lstlisting}

For Chinese text, we utilize the widely adopted Chinese segmentation tool, Jieba, for part-of-speech analysis. The core implementation code is provided below:
\begin{lstlisting}[language=python]
import jieba.posseg as psg

def extract_nouns(text):
    tagged_words = psg.cut(text)
    conditions = [x.word for x in tagged_words if 'n' in x.flag]
    return conditions
\end{lstlisting}

For Japanese text, we use the specialized Japanese morphological analysis engine, Janome, to extract nouns from the Oogiri-GO corpus. The core implementation code is as follows, where tag is \begin{CJK*}{UTF8}{goth}名詞\end{CJK*} (nouns). 
\begin{lstlisting}[language=python]
from janome.tokenizer import Tokenizer

tokenizer = Tokenizer()

def extract_nouns(text):
    tokens = tokenizer.tokenize(text)
    conditions = []
    for token in tokens:
        if tag == token.part_of_speech.split(',')[0]:
            conditions.append(token.surface)
    return conditions
\end{lstlisting}

Leveraging the three aforementioned part-of-speech analysis tools, we extract nouns from Oogiri-GO. Subsequently, we conducted deduplication on these nouns and performed manual quality screening to form the final noun set $\Sm$ for weakly-associated condition sampling in CLoT.

\clearpage
\subsection{The Details of User Study}
\label{sec:user}
\begin{figure}[ht]
  \centering
 \includegraphics[width=0.45\linewidth]{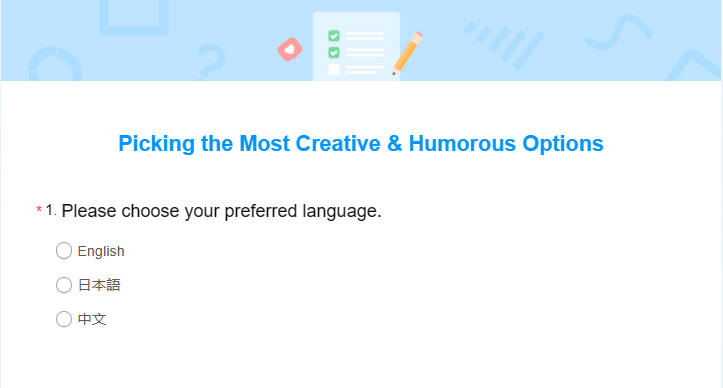}
  \caption{The questionnaire homepage of user study.}
\label{fig:homepage}
\end{figure}

\begin{figure}[ht]
  \centering
 \includegraphics[width=0.70\linewidth]{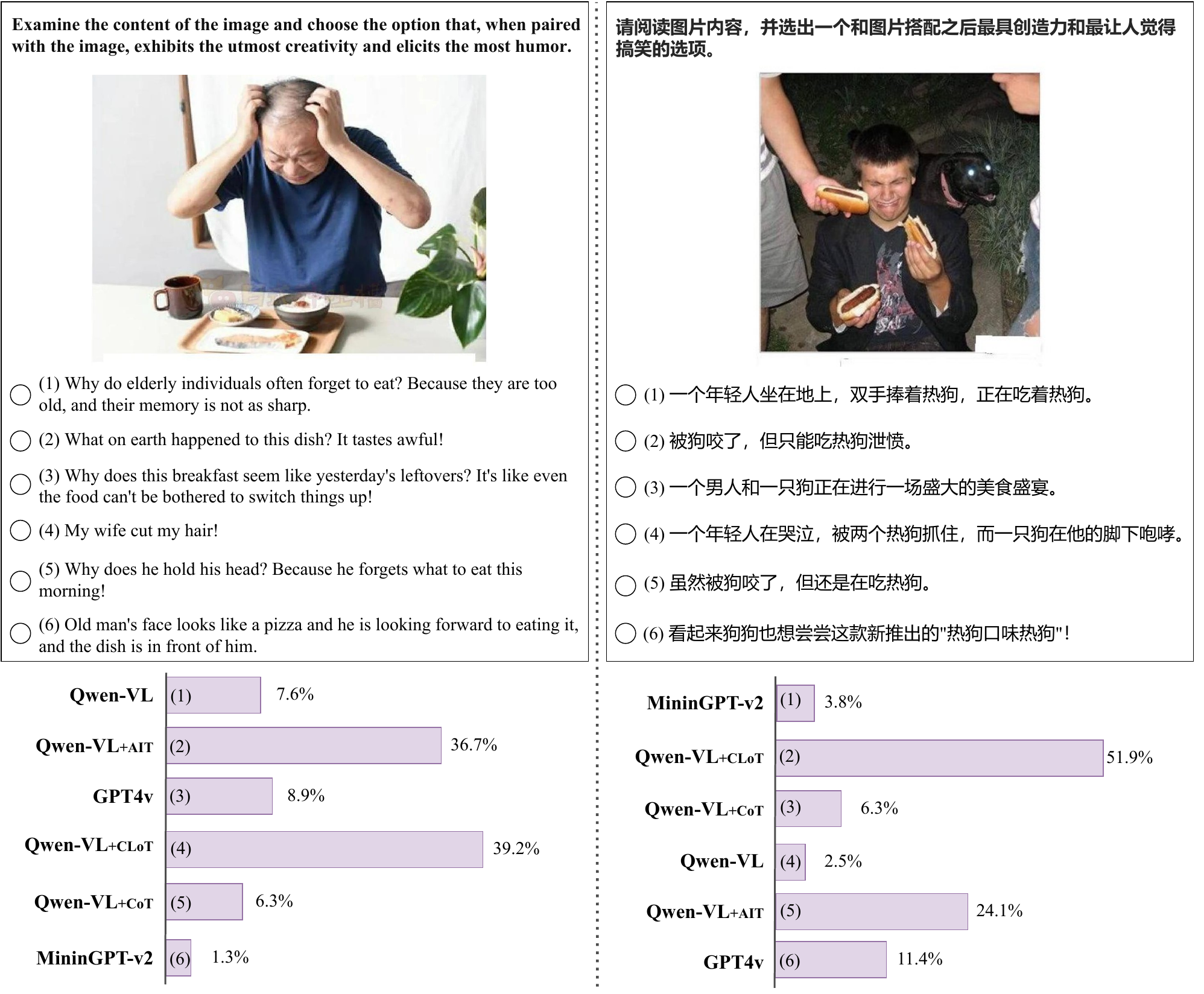}
  \caption{The questionnaire examples of user study.}
\label{fig:examples}
\end{figure}

We conduct a user preference study to directly verify the creativity of LLMs. Fig.~\ref{fig:homepage} is the questionnaire homepage of user study where users can select the preferred language of questionnaire. Subsequently, we present choice questions in the preferred language based on user selections, and ask users to choose the most creative and humorous responses as shown in Fig. \ref{fig:examples}. Here we select six advanced LLMs to generate responses for a total of eighteen questions across three tasks (IT2T, I2T and T2T), and the six responses from six distinct LLMs are randomly permuted in options.

We conduct an extensive survey through the online survey platform \footnote{https://www.wjx.cn}, ultimately collecting 154 valid questionnaires with 2772 votes. Within these collected questionnaires, we can calculate the proportion of times each LLM is selected for each question, as illustrated in Fig. \ref{fig:examples} (bottom). Finally, we aggregate the total number of times each LLM is chosen across all Oogiri samples. The ratio of this sum to the overall number of selections among all LLMs signifies the user preference for each LLM.

\clearpage

\section{The Details of other Creative Tasks}
\label{sec:othercreative}
In this section, we provide experimental details and more examples of other creative tasks, including Cloud Guessing Game (CGG), Divergent Association Task (DAT).

\subsection{The Details of Cloud Guessing Game (CGG)}

The Cloud Guessing Game (CGG) is a task that requires LLM to identify the shapes of white clouds and then select the corresponding shapes from given options. The motivation behind CGG lies in the ever-changing shapes of white clouds and creative thinking can associate specific shapes with different cloud formations, as illustrated in Fig. 9 (c-d) in the main text. Therefore, the classification performance of cloud shapes in CGG can, to some extent, analyze the LLM's LoT ability.

Specifically, the data construction process of CGG is as follows: First, we consider four categories—cat, human, and giraffe. We selected unambiguous and categorically distinct images for each category from the Internet. Taking the example of a cat, we chose the image shown in Fig.9~(a) in main text, then use Photoshop carefully to create a mask of the cat as shown in Fig.9~(b) in main text. Subsequently, employing the control diffusion model~\cite{rombach2022high,zhang2023adding,huang2023scalelong,shi2023exploring} with the following prompt, we generated a cat-like white cloud as depicted in Fig.9~(c) in main text.
\begin{mdframed}[backgroundcolor=gray!8]
\begin{minipage}{\linewidth}
(Prompt for white cloud generation) masterpiece, best quality, white cloud++, Stratus cloud,Altostratus cloud,Cirrus cloud, blue sky,light rays, sharp focus, HDR, UHD, 8K, masterpiece, Highly detailed, extreme detail detail, reality,realistic light, real, physics, reality, photo reality, Deconstruction
\end{minipage}
\end{mdframed}
\vspace{0.1cm}
Continuously generated through the control diffusion model, we manually screen until we identify 30 unambiguous and difficulty challenging white cloud images for each category. The difficulty is adjusted by the ``controlnet\_scale," a coefficient used to control the intensity of mask control. A higher value implies a stronger correlation between the generated images and masks, resulting in lower difficulty. Finally, employing the instruction template from Fig.6 (a) in the main text, we construct choice questions for 4T1, with options randomly arranged from both the ground truth category and three words sampled randomly from the unrelated word set [`chair', `cup', `sing', `jump', `rap', `basketball', `computer', `egg', `phone', `house', `lamp', `shoes'], ensuring each question's validity and clarity through manual verification. 

Finally, we constructed three choice questions for each white cloud image. The various LLMs are instructed to choose the option containing the word that best resembled the shape of the given white cloud. In the experimental setup depicted in Fig. 9 (c) of the main text, CLoT refers to the Qwen-VL+CLoT model trained as outlined in Table 2. Additionally, we used classification accuracy as a metric. The results presented in Fig. 9 (c) demonstrate that the proposed CLoT can further enhance the performance of the CGG task. This to some extent validates the versatility and effectiveness of CLoT.

\begin{figure}[h]
  \centering
 \includegraphics[width=0.99\linewidth]{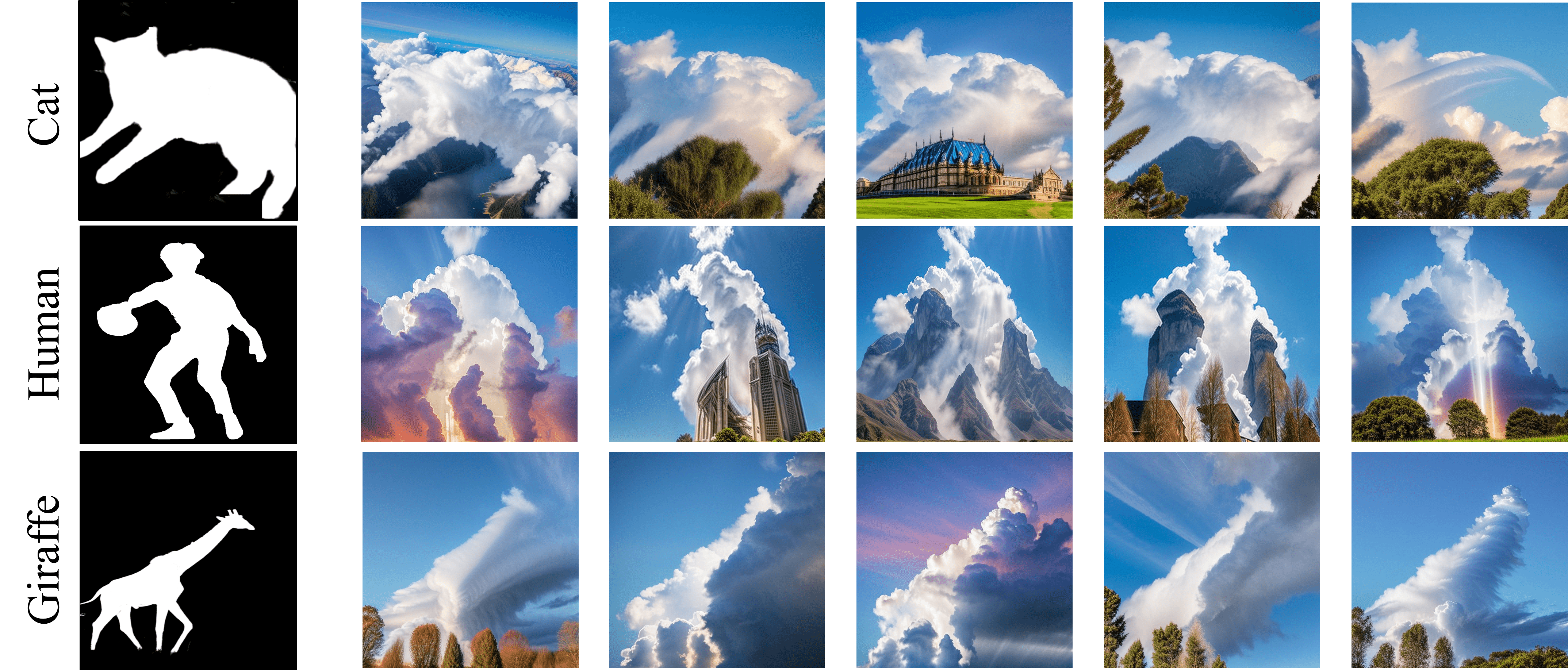}
  \caption{The examples of the generated data in CGG.}
\label{fig:morecloud}
\end{figure}

\subsection{The Details of Divergent Association Task (DAT)}

For DAT, it is a classic creativity test~\cite{beketayev2016scoring,olson2021naming} which needs participants to choose words with larger semantic distances
among 10 unrelated nouns. Building on existing research findings that suggest participants with the ability to select unrelated nouns with large semantic distances tend to have stronger associative ability, we leverage this insight to analyze the LoT ability of LLMs using the DAT benchmark~\cite{olson2021naming}. To streamline the analysis, we adapt the DAT benchmark into a series of choice questions, with the standard average semantic distance (ASD) measured by GloVe~\cite{pennington2014glove} serving as the metric. These questions challenge LLMs to choose the word from a set of nine options that differs the most from the given word.

\begin{mdframed}[backgroundcolor=gray!8]
\begin{minipage}{\linewidth}

\sethlcolor{prompt}\hl{Please carefully understand the provided question and select the option that satisfies the problem. Only one option meets the requirements. 
Question: Please select the option least relevant to the current set of words.

Words:} \sethlcolor{condition}\hl{\textless Words\textgreater}

\sethlcolor{prompt}\hl{Options: }
\sethlcolor{condition}\hl{\textless Options\textgreater}

\sethlcolor{prompt}\hl{Answer Format: Please respond in the format of 'Option id. Option content,' for example, 'A. xxx.' Response: Satisfactory option is}

\sethlcolor{response}\hl{\textless Response\textgreater}\\
\end{minipage}
\end{mdframed}
\vspace{0.1cm}

Specifically we use the instruction template above for the DAT task on LLM. Below we provide some examples of words and options:

\begin{mdframed}[backgroundcolor=gray!8]
\begin{minipage}{\linewidth}
Example 1:

 \sethlcolor{condition}\hl{\textless Words\textgreater}: Guitar Amplifier Strings Pick Melody Chord Song Musician Concert

\sethlcolor{condition}\hl{\textless Options\textgreater}: A.studio  B.hat  C.piano  D.umbrella
\end{minipage}
\end{mdframed}
\vspace{0.1cm}

\begin{mdframed}[backgroundcolor=gray!8]
\begin{minipage}{\linewidth}
Example 2:

 \sethlcolor{condition}\hl{\textless Words\textgreater}: Guitar Amplifier Strings Pick Melody Chord Song Musician Concert

\sethlcolor{condition}\hl{\textless Options\textgreater}: A.flame  B.orange  C.diamond  D.earth
\end{minipage}
\end{mdframed}
\vspace{0.1cm}

\begin{mdframed}[backgroundcolor=gray!8]
\begin{minipage}{\linewidth}
Example 3:

 \sethlcolor{condition}\hl{\textless Words\textgreater}: Soccer Amplifier Marathon Surfing Volleyball Basketball Carrot Running Yoga

\sethlcolor{condition}\hl{\textless Options\textgreater}: A.canvas  B.wire  C.volcano  D.bracelet
\end{minipage}
\end{mdframed}
\vspace{0.1cm}

\begin{mdframed}[backgroundcolor=gray!8]
\begin{minipage}{\linewidth}
Example 4:

 \sethlcolor{condition}\hl{\textless Words\textgreater}: Pepper Zucchini Eggplant Surfing Garlic Potato Carrot Bean Gymnastics

\sethlcolor{condition}\hl{\textless Options\textgreater}: A.drill  B.bee  C.hourglass  D.brick
\end{minipage}
\end{mdframed}
\vspace{0.1cm}

\begin{mdframed}[backgroundcolor=gray!8]
\begin{minipage}{\linewidth}
Example 5:

 \sethlcolor{condition}\hl{\textless Words\textgreater}: Decaf Pastry Brew Roast Forest Outdoors Compass Bean Backpack

\sethlcolor{condition}\hl{\textless Options\textgreater}: A.cake  B.whip  C.space  D.river
\end{minipage}
\end{mdframed}
\vspace{0.1cm}

\clearpage
\section{The Analysis for Self-Refinement}
\label{sec:selfrefinement}
In this section, we will further analyze why we employ a one-round self-refinement in CLoT on Oogiri-GO. Additionally, we will discuss the reasons why CLoT does not induce performance collapse during ``Explorative Self-Refinement" stage.

\subsection{The Discussion for the Round of Self-Refinement}
\label{sec:oneround}
In Section 5 of the main text, we empirically demonstrated that additional rounds do not significantly enhance the LoT ability. Consequently, we default to a one-round setting when performing self-refinement on the Oogiri-GO dataset. In this section, we delve into a detailed analysis of the underlying reasons for this phenomenon. Indeed, to achieve efficacy with multiple rounds of self-refinement, two strategies can be employed.

Firstly, by \underline{expanding diversity in creative data} (strategy 1). This involves providing the model with a more diverse set of creative data during the `` Associable Instruction Tuning" stage to enhance initial LoT capabilities. This diversity enables the LLM to continuously generate novel data effectively even under various weakly-associated conditions; 

Secondly, by \underline{ensuring diversity in the noun set $\Sm$ } (strategy 2). This implies using a more diverse and effective set of nouns in $\Sm$ for sampling weakly-associated conditions, thereby facilitating better associative capabilities in LLM and ensuring the quality of newly generated data.
\vspace{-10pt}
\begin{figure}[h]
  \centering
 \includegraphics[width=0.7\linewidth]{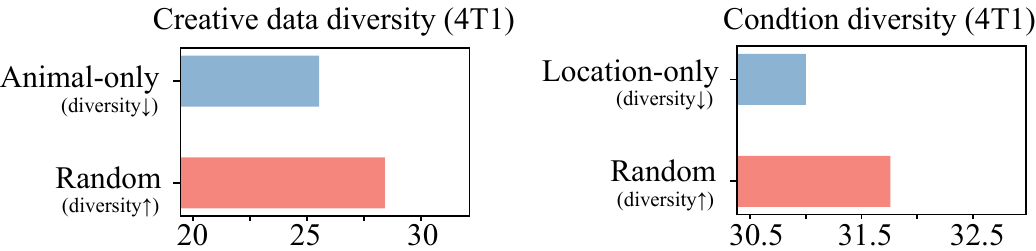}
  \caption{The impact of data diversity on CLoT performance. The baseline is Qwen-VL on I2T type Oogiri game. \textbf{Left:} For diversity of Oogiri-GO  . \textbf{Right:} For diversity of weakly-associated conditions in noun set $\Sm$.}
\label{fig:oneround}
\end{figure}

\textbf{For strategy 1.} We first substantiate the impact of creative data diversity on model performance. We conduct experiments using two subsets from the Oogiri-GO dataset: an ``animal-only" subset and a ``random" subset. The former consisted of 10,000 randomly sampled data containing animal-related nouns in responses, while the latter comprised 10,000 samples randomly drawn from the Oogiri-GO dataset. As depicted in Fig.~\ref{fig:oneround} (Left), despite having an equal data size, the diverse ``random" subset exhibited significant performance advantages. Therefore, for strategy 1, collecting a wide range of potential and diverse creative data is crucial to enhance LLM performance in generating effective new data under various weakly associated conditions. However, the inherent scarcity of high-quality creative data poses a challenge, as continuous production of innovative data by humans is not easily sustained. Additionally, the Oogiri-GO dataset already encompasses responses from a substantial portion of online Oogiri games, making it difficult to obtain a large-scale collection of new data. Hence, the inherent scarcity of innovative data constrains the further expansion of creative dataset diversity.

\textbf{For strategy 2.} Similarly, we conduct a simple experiment to illustrate the significant impact of $\Sm$ diversity on model performance. We randomly sampled 5\% of nouns from $\Sm$ to create a ``random" subset and extracted an equal number of location-related nouns to form a ``location-only" subset. The results in Fig.~\ref{fig:oneround} (Right) revealed the importance of $\Sm$ diversity for LLM's LoT ability. Therefore, for strategy 2, expanding the diversity of effective nouns in $\Sm$ is essential. 
However, when utilizing nouns from $\Sm$ for conditional generation by LLM, the number of newly generated effective nouns is limited. This limitation arises because the quantity of nouns in the response text is limited, and most nouns depend on those sampled in $\Sm$. Consequently, existing paradigms encounter difficulty in diversifying $\Sm$ due to the constraints on the number of effective nouns generated. Moreover, introducing new nouns, and potentially verbs, from external knowledge bases into $\Sm$ poses a challenge, as it necessitates ensuring that the newly added vocabulary provides sufficient clues to guide creative responses. For instance, external conditions may be entirely unrelated to the current task, making it challenging for LLM to draw upon existing knowledge for generating effective new creative data through associative thinking. The ideal scenario involves conditions that have a certain distance from the current knowledge domain but are not entirely irrelevant—termed as weakly-associated conditions.
Therefore, the intrinsic difficulty in expanding the diversity of $\Sm$ hinders the augmentation of $\Sm$ diversity.

In conclusion, \underline{\textbf{due to inherent constraints on expanding the diversity of $\Sm$ and creative data}}, existing paradigms struggle to provide sufficient diversity for multi-round self-refinement. The experiments in Section 5 of the main text indicate that one-round self-refinement effectively utilizes the existing diversity in $\Sm$ and creative data. Consequently, multiple rounds of self-refinement do not yield a significant performance boost, as one-round already achieves satisfactory performance.

\subsection{Self-Refinement doesn't Cause a Performance Collapse in CLoT}

During the training of large language models, there is a phenomenon known as ``Performance Collapse"~\cite{hataya2023contamination,shumailov2023model} while using the LLM-generated data. Specifically, due to the impressive performance of these models and their widespread use by various users, the Internet is now flooded with a vast amount of text generated by large language models, including answers, conversations, chat records, and more. Despite the generated text appearing close to those generated by human, it has irreversibly polluted Internet text data~\cite{hataya2023contamination,shumailov2023model}. This pollution is expected to result in a performance decline when large language models are retrained in the future to update parameters. The generated data from the Internet often exhibits similar patterns or implicit characteristics. Continuously feeding self-generated data to large language models~\cite{hataya2023contamination,shumailov2023model}, i.e., self-refinement, leads to an accumulation of similar data during training, restricting the diversity of model outputs and ultimately causing ``Performance Collapse".

However, for CLoT, the proposed \underline{\textbf{``Explorative Self-Refinement" stage does not lead to ``Performance Collapse"}}. This is because, (1) during this stage, the generated data is produced under the constraints of various weak-associated conditions, ensuring diversity and alleviating the issue of similar patterns; (2) in the ``Explorative Self-Refinement" stage, the generated data undergoes rigorous filtering through the discrimination ability by tuned LLM during the ``Associable Instruction Tuning" stage. This process ensures that the generated data is of high quality and mitigates the potential risk of ``Performance Collapse". These two mechanisms provide effective safeguards for CLoT when enhancing the LoT capability of LLM. Experimental results in the main text empirically demonstrate that ``Explorative Self-Refinement" does not impact model performance and significantly promotes the Leap-of-Thought ability of the model.

\clearpage
\section{Further Discussions}
\label{sec:further}
In this section, we further discuss CLoT from other perspectives.

\subsection{The Oogiri-GO Dataset doesn't Include English IT2T Type Data}
\label{sec:it2teng}

Tables 1 and 3 in the main text reveal the absence of English IT2T type data in the Oogiri-GO dataset. This can be attributed to two main reasons. 

(1) \underline{\textbf{Cultural difference.}} Firstly, Oogiri games are traditional comedy games in Japan~\cite{oogiri}, with their content heavily influenced by the cultural context of players. Currently, Oogiri games are predominantly popular in countries with similar cultures, such as China and Japan. Due to cultural differences, the more intricate IT2T type of Oogiri games is not widely embraced in English-speaking countries' online communities, where participation primarily revolves around I2T or T2T formats.

(2) \underline{\textbf{Complex processes.}} Secondly, a considerable portion of Oogiri game data on the Internet relies on bloggers and website operators who disseminate the Oogiri games through translation in their respective countries. The creation of IT2T-type Oogiri game data requires specific tools for Optical Character Recognition (OCR) \cite{memon2020handwritten,li2023trocr}, image editing \cite{perez2023poisson} and machine translation \cite{stahlberg2020neural,yang2020survey}, while other types of data are very simple to migrate to English. The Complex process for IT2T type data presents certain barriers, leading to a scarcity of IT2T content in English on the Internet.

In summary, given the rarity of English IT2T type data and the challenges associated with its creation, including the editing and translation of image text, our Oogiri-GO dataset does not encompass this type of data. Conversely, the dataset includes an ample amount of other types of data, adequately fulfilling the requirements for validating and analyzing the proposed CLoT.

\vspace{0.5cm}
\subsection{Fine-tuning Directly on Oogiri-GO is Hard to Achieve Good LoT Ability }
\label{sec:fine}
In the main text, we substantiate the efficacy of CLoT's ``Associable Instruction Tuning" and ``Explorative Self-Refinement" stages in enhancing the LoT capabilities of LLM through extensive experiments and analyses. This results in the impressive generation of humor. In this section, we revisit these two stages, asserting that a simple direct fine-tuning approach falls short in achieving sufficiently robust LoT ability.

\begin{figure}[h]
  \centering
 \includegraphics[width=0.7\linewidth]{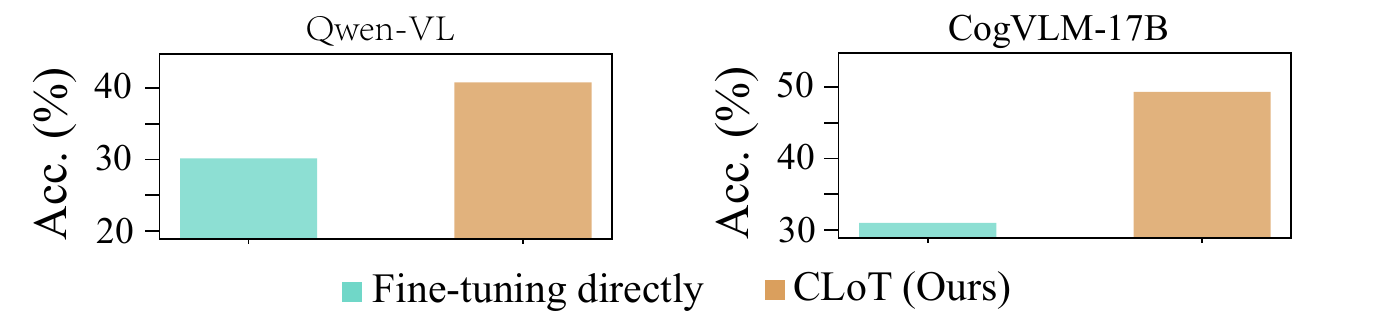}
 \vspace{-5pt}
  \caption{The performance of LLM with direct fine-tuning under 3T1 and I2T settings.}
\label{fig:onlyfine}

\end{figure}

To illustrate this conclusion, we conduct an experiment wherein the model underwent direct fine-tuning based on the instruction template of the ``Associable Instruction Tuning" stage, as outlined in the main text. Specifically, the model is fine-tuned using the template depicted in only Fig. 6 (a). The results, presented in Fig.~\ref{fig:onlyfine}, reveal a significant performance gap between direct fine-tuning on Oogiri-GO and CLoT. Two primary reasons account for this phenomenon: 

(1) \underline{\textbf{A lack of targeted associative exercises for LoT.}}  In fact, directly fine-tuning on the given creative data merely amounts to a rigorous fitting of the data. This fitting process only captures the inherent creative patterns within the data, failing to stimulate "thinking outside the box" for generating novel ideas. Furthermore, as mentioned in Appendix \ref{sec:oneround}, creative data is inherently scarce, and relying solely on dataset fitting easily leads to being trapped in local patterns. Hence, there is a pressing need for associative exercises to foster a departure from conventional thinking.

(2) \underline{\textbf{The creativity is uneven}}. While Oogiri-GO responses stem from human creativity, the creativity in these responses varies widely. Some are highly imaginative, while others are mundane. The inherent difficulty in generating creative responses, even for humans, leads to uneven quality in the dataset, with a scarcity of exceptionally high-quality creative instances. Intuitively, only such instances have the potential to stimulate the LoT capabilities of the model. Learning from general creative responses is insufficient to foster strong LoT abilities.
\vspace{-10pt}

Fortunately, CLoT not only introduces associative exercises for LLM but also leverages the varied quality of creative data, enabling LLM to discern and generate exceptionally high-quality creative responses. This distinctive approach ultimately yields performance beyond what direct fine-tuning can achieve.
 
\vspace{0.5cm}
\subsection{How to Further Enhance CLoT?}

Although CLoT has demonstrated strong efficacy in enhancing the Leap-of-Thought capability of LLM, there are still notable areas for improvement in the future. For instance, as mentioned in Appendix \ref{sec:fine}, the creative quality in data such as Oogiri-GO is uneven, and CLoT has leveraged this diversity to enhance LLM's discrimination ability towards creative data, thereby aiding in the generation of high-quality creative content. Furthermore, these human-annotated data, featuring human rankings, can be utilized to construct evaluators, which employ the innovative Reinforcement Learning from Human Feedback (RLHF) technique to further boost CLoT's performance, a pivotal approach for enhancing large language models.

Additionally, within this study, we reveal that prompting alone is insufficient to stimulate LLM's LoT ability. Despite LLM possessing rich prior knowledge and excellent reasoning capabilities, additional training is currently necessary to activate LoT. Therefore, exploring ways to maximize LoT activation through prompting or minimizing LLM training is a meaningful research direction. Moreover, the tuning of LLM's instructions, as seen in methods like LoRA, inevitably results in partial forgetting~\cite{hu2021lora,zhai2023investigating,Qwen-VL} of its inherent knowledge. For creative tasks, preserving as much of the original knowledge as possible is valuable. Hence, future work should focus on continuous learning approaches~\cite{wu2021striking,huang2021altersgd,liang2022balancing} to ensure the model retains existing knowledge to the greatest extent possible.